\documentclass[11pt]{article}
\usepackage{paralist}

\usepackage[final]{acl}

\usepackage{times}
\usepackage{latexsym}

\usepackage[T1]{fontenc}

\usepackage[utf8]{inputenc}
\usepackage{CJK}

\usepackage{microtype}

\usepackage{inconsolata}

\usepackage{graphicx}

\usepackage{soul}
\usepackage{lipsum}
\usepackage{amsmath}
\usepackage{amssymb}
\usepackage{enumitem}
\usepackage{multirow}
\usepackage{booktabs}
\usepackage[table]{xcolor}

\title{Parallel Tokenizers: Rethinking Vocabulary Design in Cross-Lingual Transfer of Low-Resource Languages}

\author{Muhammad Dehan Al Kautsar \quad Fajri Koto \\
Mohamed bin Zayed University of Artificial Intelligence\\
\texttt{\small \href{mailto:Muhammad.Dehan@mbzuai.ac.ae} {\color{black}{\{muhammad.dehan,fajri.koto\}@mbzuai.ac.ae}}}
}

\begin{document}
\maketitle
\begin{abstract}

Tokenization forms the basis of multilingual language models, yet existing methods often limit cross-lingual transfer by mapping semantically equivalent words to different embeddings. 
For example, ``\textit{I eat rice}'' in English and ``\textit{Ina cin shinkafa}'' in Hausa are typically mapped to different vocabulary indices, preventing shared representations and limiting cross-lingual generalization. This problem is even more pronounced in low-resource languages, where shared representations could offer the greatest benefit.
We introduce \textbf{parallel tokenizers}, a new framework that first trains tokenizers monolingually and then aligns their vocabularies exhaustively using bilingual dictionaries or word-to-word translation. This alignment enforces a shared semantic space across languages while naturally improving fertility balance. To assess their effectiveness, we pretrain a transformer encoder from scratch on thirteen low-resource languages and evaluate it on sentiment analysis, hate speech detection, emotion classification, and sentence embedding similarity. Across all tasks, models trained with parallel tokenizers outperform conventional multilingual baselines, confirming that rethinking tokenization is essential for advancing multilingual representation learning--especially in low-resource settings.\footnote{Code can be found at \href{https://github.com/dehanalkautsar/parallel_tokenizers}{Github}.}

\end{abstract}

\section{Introduction}

Tokenization converts text into discrete units such as words, subwords, or bytes \cite{sennrich-etal-2016-neural}. In multilingual models \cite{devlin-etal-2019-bert,conneau-etal-2020-unsupervised,xue-etal-2021-mt5}, a shared tokenizer is typically applied across languages, often disadvantaging underrepresented languages because the same semantic content is segmented into longer token sequences. This imbalance, measured by the fertility score \cite{rust-etal-2021-good}, reveals systematic inefficiencies in multilingual tokenization. While shared embeddings across languages could improve representation learning, especially for low-resource languages, current tokenizers rarely achieve such alignment.

\begin{figure}[t]
  \centering
  \includegraphics[width=0.9\columnwidth]{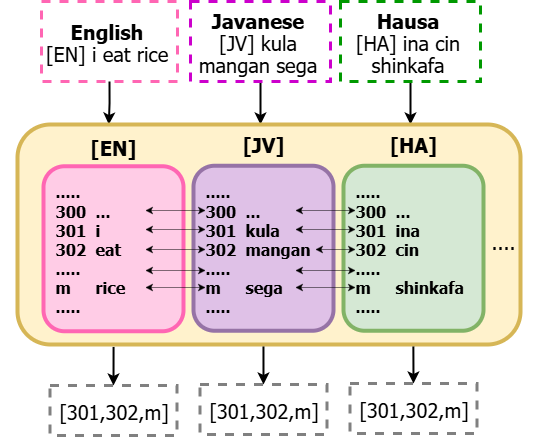}
  \caption{The overview of the parallel tokenizers. Tokens with equivalent meanings across languages are mapped to the same index and thus share the same embedding representation in the model.}
  \label{fig:intro_parallel}
\end{figure}

Beyond fertility, multilingual tokenizers also fail to align semantics across languages. Semantically equivalent words (e.g., ``\textit{eat}'' in English, ``\textit{ci}'' in Hausa, and ``\begin{CJK}{UTF8}{min}食べる\end{CJK}'' in Japanese) are mapped to entirely distinct vocabulary indices \cite{devlin-etal-2019-bert,conneau-etal-2020-unsupervised,xue-etal-2021-mt5,achiam2023gpt,bai2023qwen,dubey2024llama}, erecting artificial barriers to cross-lingual sharing.
Multilingual tokenizers also struggle with false cognates, where the same token may carry different meanings across languages. This fragmentation undermines the promise of transfer learning in multilingual models and introduces redundancy and inefficiency at the very foundation of representation learning.

Several strategies have been proposed to address these challenges, including approaches for pretraining medium- and low-resource language models \citep{kotosherkala} and byte-/character-level tokenizers \citep{xue-etal-2022-byt5,clark-etal-2022-canine,cao-2023-best}. However, semantic redundancy remains unresolved, and reducing token length disparities often comes at the cost of efficiency and semantic interpretability

\begin{figure*}[tbp]
  \centering
  \includegraphics[width=1.6\columnwidth]{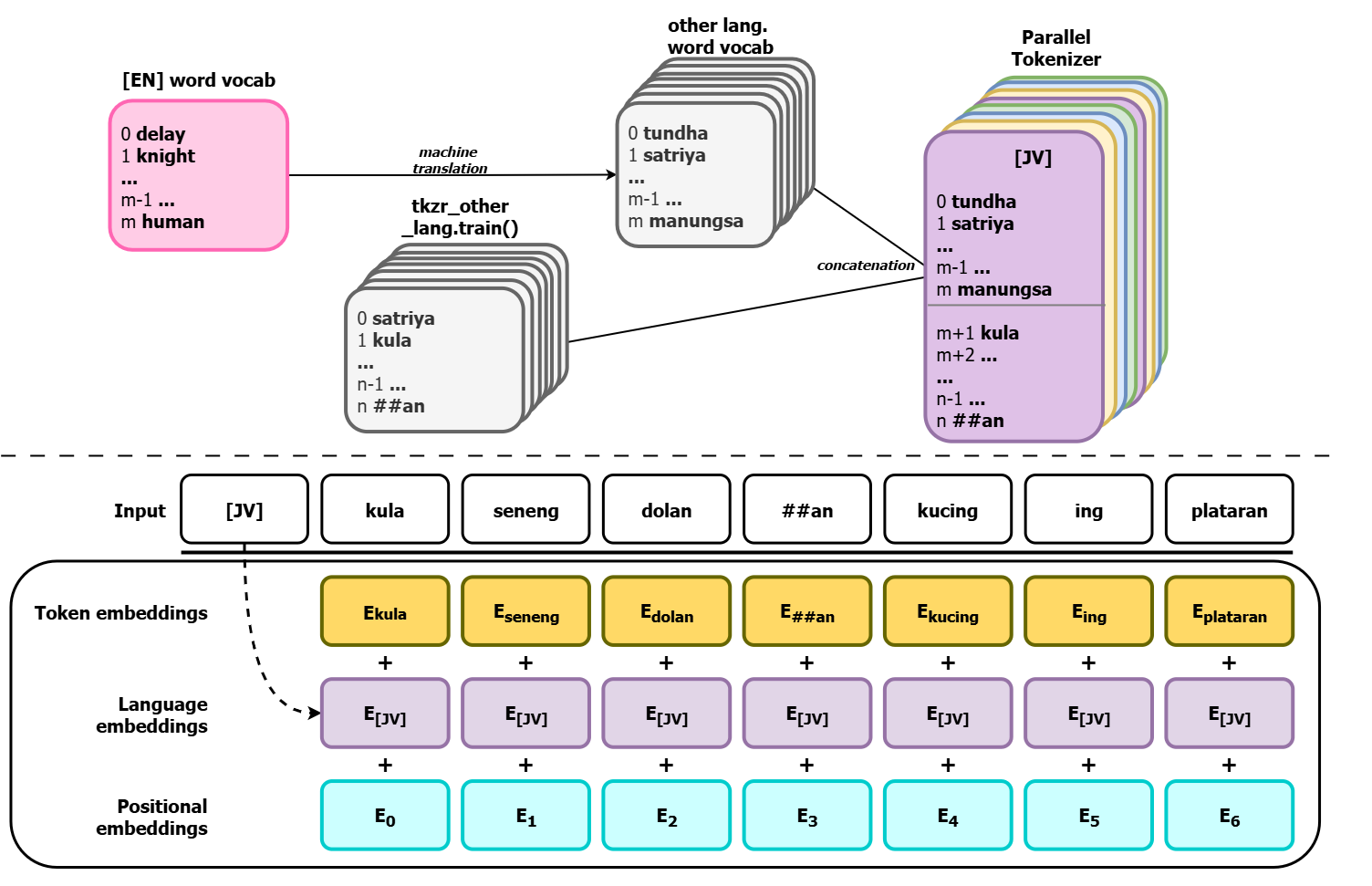}
  \caption{Design of the parallel vocabularies (top) and the model input representation (bottom). The language token (e.g., \texttt{[JV]} in the example) is not used as an explicit input token; instead, it functions as a signal to select the corresponding language identity embedding during input representation.}
  \label{fig:vocabulary_creation}
\end{figure*}

To strengthen shared representation across low-resource languages,
we propose a \textbf{parallel tokenizer} (Figure~\ref{fig:intro_parallel}). Instead of relying on a single multilingual tokenizer, 
we first train a monolingual tokenizer as a pivot and then align its vocabulary with target languages via word-level mapping, ensuring that token indices correspond to semantically equivalent words across languages.
While achieving perfect alignment is inherently difficult, we restrict alignment to exact word forms. To reduce cross-lingual interference, we further incorporate language identity embeddings (Figure~\ref{fig:vocabulary_creation}), ensuring that token representations remain unique to each language while still enabling cross-lingual semantic sharing. This design yields fairer representations, facilitates more effective transfer, and provides a scalable framework for adding new languages without retraining the entire vocabulary.

For evaluating our approach, we focus on encoder-only architectures because decoder-style generative models are considerably less suitable in low-resource settings: their generation quality degrades sharply when training data is limited. Encoder models, in contrast, emphasize semantic understanding and remain highly effective for applications such as classification and information retrieval. These properties make encoder-only architectures a clean and principled test bed for isolating the effects of our tokenization improvements, as also shown by prior works \cite{wang-etal-2022-expanding,koto-etal-2024-zero}

Our contributions can be summarized as follows:
\begin{compactitem}
    \item We propose a parallel tokenizer for multilingual language models that improves fertility balance across languages and strengthens cross-lingual transfer in encoder-only models, compared with traditional multilingual tokenizers.
    \item We demonstrate its effectiveness across multiple downstream tasks, including sentiment analysis, emotion classification, hate speech detection, and sentence embedding, all in low-resource settings.
    \item We analyze performance under varying data availability (1\%, 10\%, 50\%, 100\%), auxiliary languages, and evaluate fertility scores to measure tokenization efficiency.
\end{compactitem}

\section{Related Works}



The standard practice in multilingual language modeling is to train a single tokenizer on a combined multilingual corpus~\citep{achiam2023gpt,devlin-etal-2019-bert,xue-etal-2021-mt5,dubey2024llama,bai2023qwen}, typically using algorithms such as WordPiece~\citep{devlin-etal-2019-bert}, Byte Pair Encoding (BPE)~\citep{sennrich-etal-2016-neural}, the Unigram Language Model~\citep{kudo-2018-subword}, or SentencePiece~\citep{kudo-richardson-2018-sentencepiece}. Despite their simplicity and scalability, these approaches face the well-documented \textit{curse of multilinguality}~\citep{conneau-etal-2020-unsupervised}: monolingual tokenizers consistently outperform shared multilingual ones in fertility and segmentation quality~\citep{rust-etal-2021-good,petrov-etal-2023-language}, revealing a persistent trade-off between coverage and linguistic specialization. To mitigate these issues, prior studies have proposed balancing vocabulary allocation across languages~\citep{zheng-etal-2021-allocating,chung-etal-2020-improving,liang-etal-2023-xlm,kotosherkala}. However, while such methods improve fertility for low-resource languages, they often lead to redundant semantic representations, where tokens with equivalent meanings occupy distinct indices.

Beyond these engineering efforts, several studies have examined the conceptual role of tokenization in multilingual transfer. \citet{conneau-etal-2020-emerging} argue that parameter sharing is the primary factor enabling cross-lingual generalization. However, this line of research has not explored multilingual token embeddings in a parallel setting, where alignment across languages could better capture shared semantics. \citet{limisiewicz-etal-2023-tokenization} focus on lexical overlap, showing that surface-form similarity alone does not ensure semantic equivalence, leaving open the challenge of designing vocabularies that promote semantic alignment across languages.
Moreover, \citet{muller-etal-2021-unseen} showed that some languages exhibit limited transfer to higher-resource languages largely due to differences in the scripts used to write them. 
From a cost perspective, \citet{ahia-etal-2023-languages} found that many low-resource languages still benefit less from LLMs because of their high token fertility scores, which result in disproportionately higher computational and monetary costs. Our work seeks to narrow this gap and make LLMs more accessible and equitable for these languages.

Recent work further highlights the enduring importance of tokenization. \citet{rajaraman-etal-2024-markov} demonstrate that transformers trained on structured distributions struggle to learn effectively without discrete token boundaries, calling into question the feasibility of tokenization-free paradigms~\citep{deiseroth-etal-2024-free}. In this context, our work revisits multilingual tokenization from a new perspective, emphasizing that tokenization remains a foundational component for building semantically aligned and efficient multilingual language models.

\section{Parallel Tokenizer}

To address fertility imbalance, false cognates, and the separation of semantically equivalent words into different token indices~\citep{rust-etal-2021-good,bai2023qwen}, we introduce the Parallel Tokenizer. This section outlines (1) the construction of the parallel vocabulary and (2) the design of the model input representation.

\subsection{Parallel Vocabulary Design}
\label{sec:parallel_vocabulary}

To construct parallel vocabularies across languages, we first train an English monolingual tokenizer using WordPiece \cite{devlin-etal-2019-bert} on the English Wikipedia dump as the corpus. We set the vocabulary size to 30,522. The resulting vocabulary entries can be grouped into four categories:

\begin{enumerate}[noitemsep]
    \item \texttt{subword}, e.g., \textit{`\#\#ing'}, \textit{`\#\#able'} (28.75\% of vocabulary);
    \item \texttt{short word}, e.g., \textit{`ple'}, \textit{`szo'}, \textit{`rae'}; tokens with a length shorter than four characters (3.28\% of vocabulary);
    \item \texttt{number}, numerical tokens such as \textit{`192'}, \textit{`392'}, etc. (2.10\% of vocabulary); and
    \item \texttt{word}, which encompasses all other tokens, e.g., \textit{`care'}, \textit{`drink'}, \textit{`administration'} (65.87\% of vocabulary).
\end{enumerate}

We retain only the English \texttt{word}-type vocabulary and expand it into other languages through token-level machine translation (see Figure~\ref{fig:vocabulary_creation}). This filtering is necessary because other token types (\texttt{subword}, \texttt{short word}, and \texttt{number}) often yield unreliable or meaningless translations that rarely correspond to valid equivalents in the target languages. Using this filtered vocabulary as the base, we construct the parallel tokenizer in three steps: (1) translating the \texttt{word}-type vocabulary, (2) training a monolingual tokenizer for each target language, and (3) concatenating the resulting vocabularies.

\paragraph{Translating the \texttt{word}-type vocabulary.} We translate each English \texttt{word}-type token into its counterpart in the target language using Google Translate\footnote{\url{https://translate.google.com/}}.
 We initially considered bilingual dictionaries such as PanLex\footnote{\url{https://old.panlex.org/}},
 but found substantial limitations in both coverage and data quality, where only an average of 21.5\% of the tokens are aligned. Machine translation offers broader and more up-to-date lexical coverage. To further ensure quality, we apply back-translation to the invalid outputs, such as multiword or malformed phrases. 
 This process yields semantically aligned vocabularies that form the word-level backbone of our parallel tokenizer.

\paragraph{Training a monolingual tokenizer on the target language.}
Since the English tokenizer contains only 65.87\% \texttt{word}-type tokens, we need to expand each parallel tokenizer to the full vocabulary size of 30,522 tokens by training a monolingual tokenizer on its corresponding Wikipedia corpus. The resulting vocabularies serve as sources for additional token types. We also introduce \textit{language identity tokens} (e.g., \texttt{[JV]} in Figure~\ref{fig:vocabulary_creation}) to explicitly mark the input language and ensure consistent handling across all languages (see Section~\ref{sec:model_input_representation}).

\paragraph{Concatenating the vocabulary.}
For each target language, we concatenate two components: (1) the translated \texttt{word}-type tokens and (2) the corresponding monolingual vocabulary. We prioritize the translated \texttt{word}-type tokens, special tokens (e.g., \texttt{[SEP]}, \texttt{[UNK]}, \texttt{[CLS]})\footnote{The complete list of language tokens is provided in Appendix~\ref{sec:language_resource_availability}, Table~\ref{tab:language_resource_availability}, column `code'.}, and a 1,000-character-level subset from the English tokenizer. We then append the monolingual vocabulary, remove duplicates, and cap each final vocabulary at 30,522 entries for consistency across languages and fair comparison with the baseline. On average, across all languages defined in Section~\ref{sec:exp_setup}, we align 82\% of \texttt{word}-type tokens (slightly below 100\% due to back-translation filtering), and 61\% of all tokens are aligned in the final parallel tokenizer.

\subsection{Model Input Representation}
\label{sec:model_input_representation}

Let $x = (l, x_{1}, x_{2}, \dots, x_{i})$ denote an input sequence of length $i$, where $l \in \mathcal{L}$ represents a language identity token and $\mathcal{L}$ is the set of all such tokens (e.g., \texttt{[AM], [JV]} in Figure~\ref{fig:vocabulary_creation}), with $|\mathcal{L}|=k$. We define a collection of $k$ parallel tokenizers, each corresponding to a specific language, as
$\mathcal{T} = \{ T_{1}, T_{2}, \dots, T_{k} \}.$

Based on Figure~\ref{fig:vocabulary_creation}, the language identity token $l$ determines which tokenizer $T_{j} \in \mathcal{T}$ is used for tokenization. We then tokenize the remaining sequence $x' = (x_{1}, x_{2}, \dots, x_{i})$, excluding $l$, using $T_{j}$ to obtain the token IDs $\mathcal{I} = (I_{1}, I_{2}, \dots, I_{m})$, where $i \leq m$, along with any auxiliary representations such as attention masks and token type IDs.
The final input representation is constructed by summing the token embeddings, segment embeddings, positional embeddings, and a language identity embedding. The language identity embedding is broadcast across all $m$ tokens to provide a unique representation for each language. This additional embedding serves both as a syntactic cue and as a disambiguation signal for unaligned tokens (recall from Section~\ref{sec:parallel_vocabulary} that approximately 61\% of tokens are aligned across languages).








\section{Experimental Setup}
\label{sec:exp_setup}

\subsection{Language Setup}
\label{sec:lowreslang}
We first define the set of languages considered in this work. English is included as the base language due to its large corpus size and its role in training the base tokenizer. To assess the generalization of our approach under limited-resource conditions, we additionally select a diverse set of low-resource languages. These include Javanese (jav), Minangkabau (min), Sundanese (sun), Swahili (swa), Acehnese (ace), Amharic (amh), Balinese (ban), Hausa (hau), Igbo (ibo), Kinyarwanda (kin), Oromo (orm), Tigrinya (tir), and Twi (twi), all of which are considered low-resource languages.

Most of these languages use the Latin script, while only Amharic and Tigrinya use the Amharic script. The selection is guided by the availability of essential resources, including Wikipedia dumps\footnote{\url{https://dumps.wikimedia.org/}. Resources were downloaded in November 2024.} for tokenizer and transformer pretraining, Google Translate for constructing parallel tokenizers, and FLORES+~\cite{costa2022no} for evaluating tokenization metrics.

\subsection{Tokenizers \& Models}


To fairly compare tokenization strategies, we train two encoder-only models from scratch with identical pretraining setups but different tokenizers. The first uses a single shared tokenizer trained on the combined Wikipedia data of 13 languages, representing the standard multilingual approach; we refer to this setting as \textit{Single-13L}. The second applies our proposed parallel tokenization method (\textit{Parallel-13L}) trained on the same data. In both cases, model parameters are randomly initialized to isolate the effect of tokenization on multilingual representation learning.

\subsection{Evaluation Setup}

To evaluate the effectiveness of our proposed parallel tokenizer, we conduct three main experiments in low-resource languages: (i) tokenization analysis, (ii) sequence classification, and (iii) cross-lingual representation similarity.

\paragraph{Tokenization Analysis.}
We use the FLORES+ dataset~\citep{costa2022no, wmt24-4african}\footnote{\url{https://huggingface.co/datasets/openlanguagedata/flores_plus}}, a parallel corpus covering 200 languages, to evaluate the effectiveness of our parallel tokenizer against both multilingual and monolingual tokenizers in 1,012 parallel texts. All 13 languages studied in our experiments are represented in FLORES+. We report two metrics: (i) Fertility score: the average number of tokens per word, indicating how compact or fragmented the tokenization is; and (ii) Parity score~\citep{petrov-etal-2023-language}: a cross-lingual metric that compares token counts across two languages,\footnote{We select Hausa as the reference language for comparison, as it provides the largest available resource during pretraining.} assigning the best score when the counts are identical.

\paragraph{Sequence Classification.}
We conduct experiments on sentiment analysis, hate speech detection, and emotion classification in low-resource languages by fine-tuning the language model jointly on data from all languages. To analyze cross-lingual transfer, we vary the training size at 1\%, 10\%, 50\%, and 100\% of the available data. For sentiment analysis and hate speech detection, we use the NusaX-Senti~\citep{winata-etal-2023-nusax} and AfriHate~\citep{muhammad-etal-2025-afrihate} datasets, respectively, and evaluate both using macro-F1 over three multi-classes. For emotion classification, we use EthioEmo~\citep{belay-etal-2025-ethioemo} and BRIGHTER~\citep{muhammad-etal-2025-brighter}, evaluated with weighted F1 for consistency across six multi-label tasks. A detailed overview of the dataset languages is provided in Table~\ref{tab:benchmark_language_coverage} (Appendix~\ref{sec:benchmark_language_coverage}).

\paragraph{Cross-Lingual Representation Similarity.}
We perform bitext mining~\citep{artetxe-schwenk-2019-margin} on FLORES+ using the contextualized representations from the final layer of the language models. Following the setup of~\citet{huang-etal-2025-modular}, we evaluate \textbf{all possible language pairs} covered by our models. Performance is measured using the error rate of the \texttt{xsim} score~\citep{artetxe-schwenk-2019-margin}. Additionally, we apply Principal Component Analysis (PCA)~\citep{Jolliffe2002} to visualize cross-lingual clustering, comparing our parallel tokenizer with the multilingual model that uses a single tokenizer. Representations are considered semantically aligned when sentences from different languages cluster together, rather than being separated by language identity.

\subsection{Hyperparameters and Resources}

We use an NVIDIA RTX A6000 GPU with 48GB of VRAM for both pretraining and fine-tuning. Pretraining follows the masked language modeling (MLM) setup using a learning rate of 5e-5 for the \textit{Single-13L} model and 1e-4 for our proposed \textit{Parallel-13L} model. Training is performed on 394M tokens per epoch (6M tokens for the development set) for a total of 122,850 steps or 50 epochs. The value is determined empirically based on the convergence of the pretraining loss on the development set.

For evaluation, we set the batch size to 20, the maximum sequence length to 128, and the maximum number of epochs to 100, with early stopping triggered after 5, 2, 3, and 3 epochs for NusaX-senti, AfriHate, EthioEmo, and BRIGHTER, respectively. Learning rates were adopted from the original benchmark recommendations: 1e-5 for NusaX-senti and BRIGHTER, and 5e-5 for AfriHate and EthioEmo. Each experiment was repeated three times with different seeds (1, 12, and 123), and we report the mean performance together with the corresponding standard deviations.

\section{Results and Analysis}
\label{sec:results_section}

\subsection{Tokenization Qualities}

\begin{table*}[t]
\centering
\resizebox{0.85\textwidth}{!}{%
\small
\begin{tabular}{l|cc|c|cc|c}
\hline
\multirow{3}{*}{\textbf{lang.}} & \multicolumn{3}{c|}{\textbf{tokens/word (fertility score) $\downarrow$}} & \multicolumn{3}{c}{\textbf{parity score $\downarrow$}} \\
\cline{2-7}
 & \multicolumn{2}{c|}{\textbf{Multilingual Tokenizer}} & \textbf{Mono-} & \multicolumn{2}{c|}{\textbf{Multilingual Tokenizer}} & \textbf{Mono-} \\
 \cline{2-3} \cline{5-6}
 & \textbf{Single-13L} & \textbf{Parallel-13L (ours)} & \textbf{lingual} & \textbf{Single-13L} & \textbf{Parallel-13L (ours)} & \textbf{lingual} \\
\hline
jav & 
 1.65 & \textbf{1.48} & 1.43 & \textbf{1.06} & 1.13 & 1.69 \\
min & 
 1.79 & \textbf{1.50} & 1.46 &  \textbf{1.05} & 1.09 & 1.68 \\
sun & 
 1.74 & \textbf{1.50} & 1.44 &  \textbf{1.01} & 1.13 & 1.74  \\
swa & 
 1.65 & \textbf{1.42} & 1.38 &  \textbf{1.01} & 1.10 & 1.63 \\
ace & 
 1.95 & \textbf{1.54} & 1.61  & 1.20 & 1.02 & 1.51 \\
amh & 
 2.43 & \textbf{1.82} & 1.66 &  1.21 & \textbf{1.06} & 2.33 \\
ban & 
 1.77 & \textbf{1.53} & 1.48 &  \textbf{1.03} & 1.07 & 1.63 \\
hau & 
 1.38 & \textbf{1.33} & 1.30 &  1.00 & 1.00 & 1.00 \\
ibo & 
 1.54 & \textbf{1.50} & 1.48 &  \textbf{1.10} & 1.12 & 1.32 \\
kin & 
 2.10 & \textbf{1.70} & 1.63 &  1.21 & \textbf{1.03} & 1.53 \\
orm & 
 2.40 & \textbf{1.82} & 1.72 &  1.35 & \textbf{1.07} & 1.55 \\
tir & 
 2.61 & \textbf{1.83} & 1.85 &  1.45 & \textbf{1.06} & 2.14 \\
twi & 
 1.55 & \textbf{1.38} & 1.35 &  1.16 & \textbf{1.07} & 1.42 \\
\hline
\textit{avg} & 
 1.89 & \textbf{1.57 (\textcolor{blue}{$\downarrow$0.32})} & 1.52 &  1.14 & \textbf{1.07 (\textcolor{blue}{$\downarrow$0.07})} & 1.63 \\
\hline
\end{tabular}}
\caption{Fertility and parity scores for each tokenizer across all languages. The \textbf{bold} scores indicate the best performance in multilingual tokenizers. Monolingual tokenizers are also included to provide a comparative perspective between monolingual and multilingual tokenizers.}
\label{tab:tokenization_analysis}
\end{table*}

\begin{table*}[tbp]
\centering
\resizebox{0.85\textwidth}{!}{%
\begin{tabular}{c|c|c|c|c|c|c}
\hline
\textbf{\#training} & \textbf{Multilingual} & \textbf{\textit{Sentiment}} & \textbf{\textit{Hate Speech}} & \multicolumn{2}{c|}{\textit{\textbf{Emotion Classification}}} & \multirow{2}{*}{\textbf{Avg $\uparrow$}} \\
\cline{3-6}
 \textbf{data} & \textbf{Tokenizer} & \textbf{NusaX-senti} $\uparrow$ & \textbf{AfriHate} $\uparrow$ & \textbf{EthioEmo} $\uparrow$ & \textbf{BRIGHTER} $\uparrow$ & \\
\hline
\multirow{2}{*}{100\%} 
 & Single-13L      & 76.09 \textsubscript{($\pm$1.27)} & 69.61 \textsubscript{($\pm$1.19)} & 54.98 \textsubscript{($\pm$0.98)} & 48.26 \textsubscript{($\pm$0.94)} & 62.24\\
 & Parallel-13L (ours)   &  \textbf{76.16 \textsubscript{($\pm$1.05)}} &  \textbf{69.80 \textsubscript{($\pm$1.20)}} &  \textbf{57.01* \textsubscript{($\pm$0.84)}} &  \textbf{49.68* \textsubscript{($\pm$1.89)}} & \textbf{63.16} \\
\hline
\multirow{2}{*}{50\%} 
 & Single-13L      & 72.47 \textsubscript{($\pm$1.36)} & 67.26 \textsubscript{($\pm$1.36)} & 51.33 \textsubscript{($\pm$1.48)} & 45.77 \textsubscript{($\pm$1.46)} & 59.21 \\
 & Parallel-13L (ours)   &  \textbf{73.52* \textsubscript{($\pm$1.17)}} &  \textbf{67.40 \textsubscript{($\pm$0.94)}} &  \textbf{54.27* \textsubscript{($\pm$1.78)}} &  \textbf{46.76* \textsubscript{($\pm$1.51)}} & \textbf{60.49} \\
\hline
\multirow{2}{*}{10\%} 
 & Single-13L      & 64.76 \textsubscript{($\pm$1.84)} & 59.80 \textsubscript{($\pm$2.09)} &  \textbf{42.10 \textsubscript{($\pm$2.19)}} & 36.64 \textsubscript{($\pm$2.23)} &  50.83 \\
 & Parallel-13L (ours)   &  \textbf{66.16* \textsubscript{($\pm$2.00)}} &  \textbf{60.17 \textsubscript{($\pm$1.50)}} & 41.35 \textsubscript{($\pm$2.21)} &  \textbf{38.51 \textsubscript{($\pm$1.27)}} & \textbf{51.55} \\
\hline
\multirow{2}{*}{1\%} 
 & Single-13L      & 31.54 \textsubscript{($\pm$3.87)} & 44.70 \textsubscript{($\pm$4.45)} &  \textbf{27.37 \textsubscript{($\pm$1.88)}} & 17.54 \textsubscript{($\pm$2.57)} & 30.29 \\
 & Parallel-13L (ours)   &  \textbf{33.83 \textsubscript{($\pm$3.90)}} &  \textbf{46.73 \textsubscript{($\pm$2.52)}} & 26.58 \textsubscript{($\pm$1.37)} &  \textbf{18.88 \textsubscript{($\pm$3.96)}} & \textbf{31.51} \\
\hline
\end{tabular}
}
\caption{Performance comparison of tokenizer setups on sentiment analysis, hate speech detection, and emotion classification tasks. Best scores per benchmark setup are highlighted in \textbf{bold}. * indicates a statistically significant result (paired two-tailed t-tests), while $\pm$ represents the standard deviation across seeds.}
\label{tab:results}
\end{table*}


Table~\ref{tab:tokenization_analysis} compares our \textit{Parallel-13L} tokenizer with \textit{Single-13L} and the monolingual tokenizers. In terms of fertility score, our method substantially and consistently outperforms the multilingual baseline, exceeding by an average of 0.32 points, indicating more compact segmentation (i.e., fewer tokens per word). The only slight gap (0.05) appears against the monolingual tokenizer, which is expected since monolingual models are optimized for their respective languages. This highlights the effectiveness of our approach in reducing the number of tokens during multilingual tokenization, thereby lowering computational requirements.

For the parity score, which quantifies consistency in tokenization across parallel texts and reflects cross-lingual alignment, \textit{Parallel-13L} achieves the best overall performance, outperforming the baseline (\textit{Single-13L}) by 0.07 points on average. The \textit{Single-13L} performs notably worse on Amharic-script languages (Amharic and Tigrinya) due to script-specific segmentation differences, while monolingual tokenizers score lowest by design, as parity inherently measures multilingual generalization.



\subsection{Sequence Classification}
Across three seeds, Table~\ref{tab:results} shows that our parallel tokenizer (\textit{Parallel-13L}) consistently outperforms the baseline across training sizes and evaluation metrics. On average, \textit{Parallel-13L} exceeds the baseline by 0.92\%, 1.28\%, 0.72\%, and 1.22\% F1 at the 100\%, 50\%, 10\%, and 1\% training data levels, respectively. These results demonstrate the effectiveness of a parallelized tokenizer design in enhancing cross-lingual transfer, particularly in low-resource scenarios.

The improvements in our approach are consistent across the NusaX-senti, AfriHate, and BRIGHTER datasets, where \textit{Parallel-13L} outperforms the \textit{Single-13L} baseline regardless of training size. The only exception occurs with EthioEmo, where \textit{Single-13L} achieves higher scores than \textit{Parallel-13L} only at the 10\% and 1\% settings. Detailed per-language benchmark results are provided in Appendix~\ref{sec:apdx_main_result_detail} (Tables~\ref{tab:results_nusaxsenti_scratch}, \ref{tab:results_afrihate_scratch}, \ref{tab:results_ethioemo_scratch}, and \ref{tab:results_brighter_scratch} for NusaX-senti, AfriHate, EthioEmo, and BRIGHTER, respectively).

\subsection{Cross-Lingual Representation Similarity}

\begin{figure}[t]
  \centering
  \includegraphics[width=0.45\linewidth]{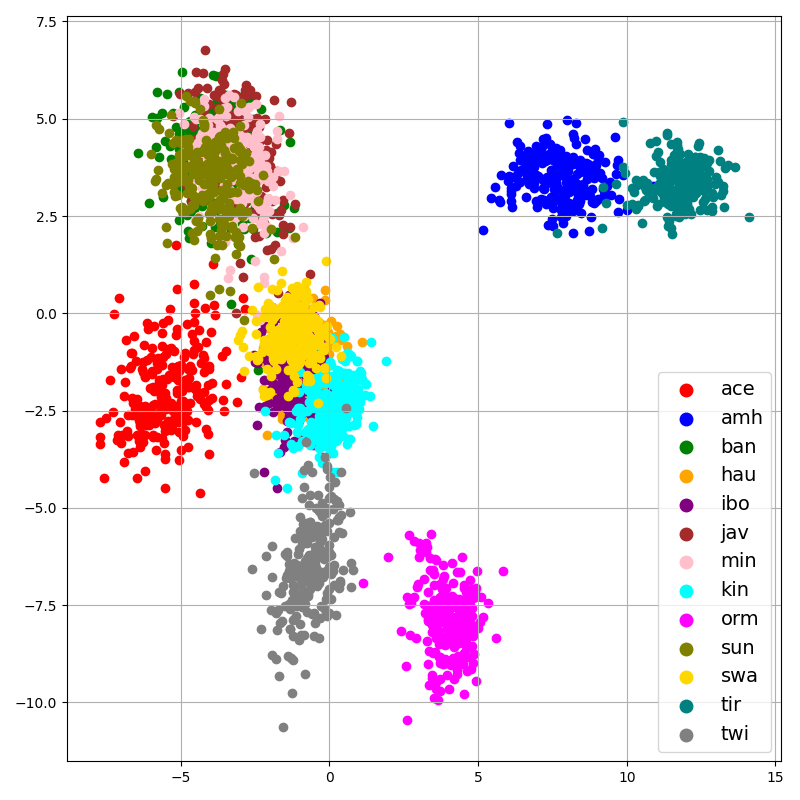} \hfill
  \includegraphics[width=0.45\linewidth]{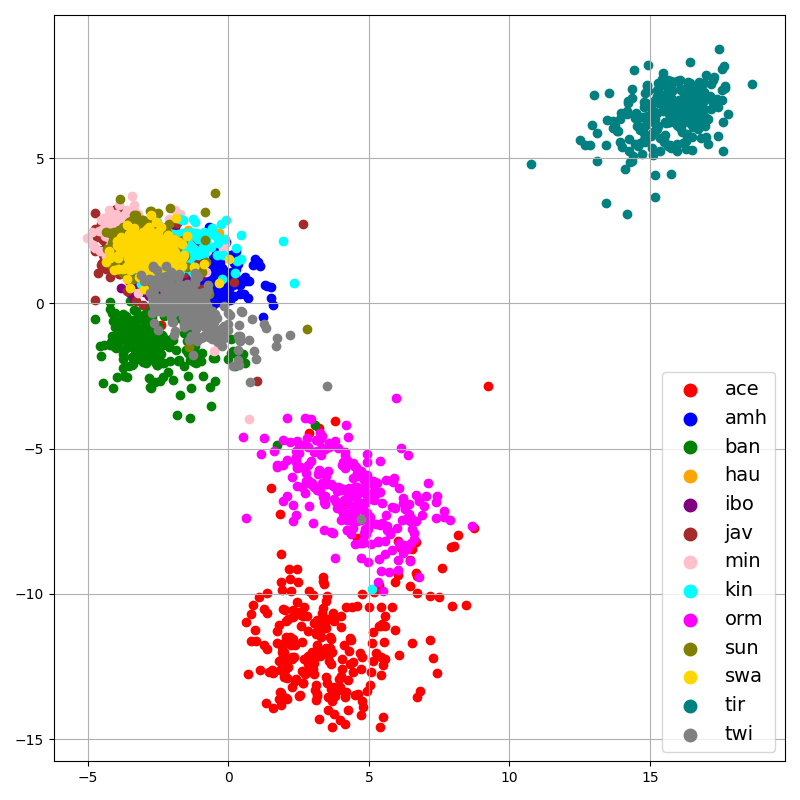}
  \caption {PCA visualization of the last hidden states from \textit{Single-13L} (left) and \textit{Parallel-13L} (right) models on the FLORES+ dataset.}
  \label{fig:pca_flores}
\end{figure}

Using the parallel FLORES+ corpus, we visualize cross-lingual representation similarity for each model by extracting the final hidden states of 250 sampled parallel sentences. If a model learns shared representations across languages, PCA should reveal clusters based on semantic similarity rather than language identity. Figure~\ref{fig:pca_flores} compares \textit{Single-13L} and \textit{Parallel-13L} in 2D space. \textit{Single-13L} tends to cluster representations by language family, for example, Indonesian languages (Sundanese, Minangkabau, Balinese, and Javanese) group together, while African languages form a separate cluster. In contrast, \textit{Parallel-13L} yields more compact cross-lingual clusters, indicating stronger semantic alignment, with only Acehnese, Oromo, and Tigrinya appearing as outliers for both due to limited data.

We further evaluate cross-lingual similarity through bitext mining. As shown in Table~\ref{tab:bitext_mining_overall}, \textit{Parallel-13L} achieves the lowest \texttt{xsim} error rate and the highest number of best scores across language pairs (details in Appendix~\ref{sec:bitext_mining_details}). These results confirm that \textit{Parallel-13L} learns more semantically consistent representations across languages than \textit{Single-13L}, reinforcing its advantage for cross-lingual tasks.

\begin{table}[t]
\centering
\tabcolsep=0.1cm
\resizebox{0.9\linewidth}{!}{%
\begin{tabular}{l|cc}
\hline
& \textbf{Single-13L} & \textbf{Parallel-13L (ours)} \\
\hline
Avg. err. \texttt{xsim} $\downarrow$ &  83.56 & \textbf{74.08} \\ 
\# of best \texttt{xsim} $\uparrow$ &  12 & \textbf{66} \\
\hline
\end{tabular}
}
\caption{Bitext mining metric meta-scores on pretrained models under different setups.}
\label{tab:bitext_mining_overall}
\end{table}

\subsection{Cross-Lingual Transfer under Limited Target-Language Data}

\begin{table}[t]
\centering
\resizebox{0.9\linewidth}{!}{%
\begin{tabular}{c|l|cc}
\hline
\multicolumn{2}{c|}{\textbf{\#data in tgt language}} & \textbf{Single-13L} & \textbf{Parallel-13L (ours)} \\
\hline
\multirow{4}{*}{50\%} 
& NusaX-senti  &  75.22 &  \textbf{75.42} \\ 
& AfriHate  & \textbf{67.06} & 67.00 \\
& EthioEmo  &  51.73 &  \textbf{53.63} \\
& BRIGHTER  &  45.00 &  \textbf{47.11} \\
\hline
\multirow{4}{*}{0\%} 
& NusaX-senti  &  71.83 &  \textbf{72.70} \\
& AfriHate  &  37.27 &  \textbf{40.67} \\
& EthioEmo  &  \textbf{23.75} & 23.20 \\
& BRIGHTER  &  15.04 &  \textbf{19.73} \\
\hline
\end{tabular}}
\caption{Benchmarking results under limited target-language data.}
\label{tab:results_overall_crosslingual_scratch}
\end{table}

While our main experiments utilize data from all languages during training, we also analyze the model's behavior when the target-language data is scarce or unavailable--a common scenario in low-resource settings. To simulate this condition, we exclude the target language from the training corpus and evaluate two setups: (i) 0\% target-language data (zero-shot) and (ii) 50\% target-language data (partial-shot).
Table~\ref{tab:results_overall_crosslingual_scratch} shows that our method consistently achieves the highest F1 scores in both settings. The only exceptions are a slight decrease of 0.06\% on AfriHate with 50\% target data and 0.55\% on EthioEmo with 0\% target train data. Overall, \textit{Parallel-13L} outperforms the baseline, demonstrating stronger cross-lingual transfer than the traditional single-shared tokenizer. The detailed results are presented in Appendix~\ref{sec:cross_lingual_auxiliary} in Table~\ref{tab:results_nusaxsenti_crosslingual_scratch}, \ref{tab:results_afrihate_crosslingual_scratch}, \ref{tab:results_ethioemo_crosslingual_scratch}, and \ref{tab:results_brigher_crosslingual_scratch}.

\subsection{Continual Pre-Training Experiments}
\label{sec:cpt}

\begin{table}[t]
\centering
\small
\resizebox{0.45\textwidth}{!}{%
\begin{tabular}{l|ccc}
\hline
 & \textbf{mBERT} & \textbf{S-13L}* & \textbf{P-13L}* \textbf{(ours)} \\
\hline
\rowcolor{yellow!50} \multicolumn{4}{c}{\textbf{Sequence Classification Benchmark}} \\
NusaX-senti & 73.1 \textsubscript{($\pm$1.7)} & 78.3 \textsubscript{($\pm$1.0)} & \textbf{78.6} \textsubscript{($\pm$1.2)} \\ 
Afrihate    & 62.8 \textsubscript{($\pm$2.4)} & \textbf{71.3} \textsubscript{($\pm$1.9)} & 70.1 \textsubscript{($\pm$0.9)} \\
EthioEmo    & 30.1 \textsubscript{($\pm$7.0)} & 56.7 \textsubscript{($\pm$0.9)} & \textbf{57.4} \textsubscript{($\pm$0.8)}\\
BRIGHTER    & 48.5 \textsubscript{($\pm$1.3)} & \textbf{50.5} \textsubscript{($\pm$0.9)} & 50.3 \textsubscript{($\pm$1.2)} \\
\hline
\rowcolor{blue!20} \multicolumn{4}{c}{\textbf{Bitext Mining Benchmark}} \\
Avg. \texttt{xsim} $\downarrow$ & 91.33 & 72.18 & \textbf{69.34} \\ 
\# of best \texttt{xsim} $\uparrow$ & 1 & 29 & \textbf{48} \\
\hline
\end{tabular}
}
\caption{Benchmarking results under the continual pretraining setting (top: sequence classification benchmark; bottom: bitext mining benchmark). S-13L* and P-13L* denote \textit{Single-13L} and \textit{Parallel-13L}, respectively, both of which are continually pretrained from the mBERT model.}
\label{tab:cpt_vs_scratch}
\end{table}

We perform continual pretraining (CPT)\footnote{i.e., extending the original pretraining objective (MLM) on additional data.} from mBERT~\cite{devlin-etal-2019-bert} by adapting its vocabulary and tokenizer to our parallel setup. Newly introduced token embeddings are initialized by averaging the mBERT embeddings of their constituent subwords, as determined by the original mBERT tokenizer~\cite{koto-etal-2021-indobertweet}. Unlike pretraining from scratch, this approach retains all mBERT parameters while updating the token embedding layer to accommodate the new vocabulary. In \textit{Single-13L}, each token in the new vocabulary is mapped to its corresponding English subwords in mBERT, and its embedding is initialized by averaging the associated subword embeddings. In \textit{Parallel-13L}, each token is tokenized into subwords across all 13 languages, and its embedding is initialized by averaging the pooled subword embeddings across languages.

Table~\ref{tab:cpt_vs_scratch} shows that \textit{Single-13L} and \textit{Parallel-13L} achieve similar sequence classification performance under continual pretraining (CPT), and both outperform the original mBERT when it is fine-tuned directly without CPT. However, our method yields stronger cross-lingual alignment (see Figure~\ref{fig:pca_flores_cpt} in Appendix~\ref{sec:apdx_cpt_representation_similarity}), and bitext mining benchmarking confirms more consistent representations than the baselines. This indicates that even when task scores converge, \textit{Parallel-13L} maintains a structural advantage in preserving cross-lingual coherence.

\subsection{Unseen Languages Performance in Downstream Task}
\label{sec:impact_lang_id}

We further analyze the role of language identity within the parallel tokenizer framework, particularly for unseen languages. This evaluation is motivated by practical scenarios where the input language may not be included in the multilingual training set, or when an incorrect language identity is inadvertently supplied at inference time. In such cases, the parallel tokenizer is forced to operate with a mismatched tokenizer-language identity pairing, allowing us to assess how robustly the model handles mis-specified or unseen language contexts.


To examine this effect, we design an extreme evaluation scenario using \textbf{five unseen languages} from the NusaX-senti dataset: bbc (Toba Batak), bjn (Banjarnese), bug (Buginese), mad (Maduranese), and nij (Ngaju). It is worth noting that this scenario is rare in practice, since the target language is typically known when applying multilingual models. Moreover, such a scenario is even less likely given that language identification systems are highly accurate~\cite{adebara-etal-2022-afrolid}. We evaluate \textit{Single-13L} and \textit{Parallel-13L} under two training data regimes: 100\% and 1\% of the available training data.

Table~\ref{tab:nusax_unseen} shows that when \textit{Parallel-13L} is forced to use an incorrect language identity, \textit{Single-13L} performs better in the full-data (100\%) setting. However, in the 1\% setting, which better reflects realistic unseen and/or low-resource language scenarios, the parallel tokenizer demonstrates clear advantages, even when the tokenizer is selected based on geographical proximity (Geo) or averaged across all tokenizers (Avg). This highlights a trade-off between stability in high-resource settings and robustness in low-resource scenarios; while such extreme mismatches are rare in practice, our method shows superior generalization when training data is severely limited.

\begin{table}[t]
\centering
\small
\tabcolsep=3.5pt
\resizebox{0.45\textwidth}{!}{%
\begin{tabular}{l|l|ccccc|c}
\toprule
\textbf{\#training} & \textbf{tokenizer} &\textbf{ bbc} & \textbf{bjn }& \textbf{bug} & \textbf{mad} & \textbf{nij} & \textbf{avg} \\
\midrule
\multirow{4}{*}{100\%} & S-13L & \textbf{52.87} & \textbf{69.00} & \textbf{51.08} & 61.76 & 59.92 & \textbf{58.93} \\
 & P-13L (Best) & 51.67 & 65.50 & 48.80 & \textbf{61.81} & \textbf{60.72} & 57.70 \\
 & P-13L (Geo) & 51.67 & 65.50 & 40.26 & 59.67 & 53.71 & 54.16 \\
 & P-13L (Avg) & 46.73 & 62.13 & 44.29 & 60.64 & 54.62 & 53.68 \\
 \midrule
\multirow{4}{*}{1\%} & S-13L & 26.70 & 28.72 & 29.10 & 28.38 & 30.36 & 28.65 \\
 & P-13L (Best) & \textbf{39.37} & \textbf{36.63} & \textbf{38.91} & \textbf{35.54} & \textbf{38.37} & \textbf{37.76} \\
 & P-13L (Geo) & 31.56 & 34.06 & 26.55 & 33.59 & 30.26 & 31.20 \\
 & P-13L (Avg) & 34.73 & 32.59 & 30.13 & 33.72 & 31.69 & 32.57 \\
\bottomrule
\end{tabular}}
\caption{Benchmarking results for unseen languages in NusaX-senti. S-13L and P-13L denote \textit{Single-13L} and \textit{Parallel-13L}, respectively. For the parallel setting, `Best' denotes the best-performing tokenizer pair, `Geo' selects the tokenizer based on the geographically closest language, and `Avg' reports the average performance obtained using all tokenizers for that language.}
\label{tab:nusax_unseen}
\end{table}

\section{Discussion}
\label{sec:discussion}

As described in Section~\ref{sec:parallel_vocabulary}, we use machine translation (MT) via the Google Translate API to translate English vocabulary into several low-resource languages. Because the system is primarily designed for sentence-level translation, we evaluate its word-level performance to ensure proper alignment with the English vocabulary. To examine this, we sample 100 vocabulary items each from Minangkabau (\texttt{min}), Javanese (\texttt{jav}), and Sundanese (\texttt{sun}), and ask fluent annotators to assess and mark whether the translations are correct or not. The average accuracy is 83\% (details: \texttt{min}=76\%, \texttt{jav}=84\%, \texttt{sun}=89\%). While the alignment is not perfect, the MT-based approach achieves substantial scores, given that the languages themselves are low-resource.

The proposed parallel tokenizers also offer advantages in handling interlingual homographs (e.g., ``burro'', which means ``donkey'' in Spanish but ``butter'' in Italian), as these ``burro''s are assigned different indices across languages. This contrasts with standard tokenization schemes, which may conflate such cases.

Nevertheless, even when the parallel tokenizer produces multiple subword units instead of a single full-form token, our approach still offers clear advantages over traditional multilingual tokenizers. Because we attach a language embedding to every token, each subword retains a language-specific identity, eliminating cross-lingual semantic ambiguity. In contrast, standard multilingual tokenizers reuse identical subword pieces across languages, regardless of whether they share meaning, often conflating unrelated semantics. Importantly, our fertility scores empirically confirm that the proposed tokenizer yields more compact and semantically faithful segmentations, and these improvements translate into consistent performance gains across downstream tasks.




\section{Conclusion \& Future Works}

We introduce parallel tokenizers, constructed by first training a monolingual tokenizer as a pivot and then aligning its word vocabulary with other languages using machine translation, ensuring that semantically equivalent words across languages share the same representation. We evaluate this approach along three dimensions. In terms of tokenization quality, it yields the lowest fertility and parity scores, indicating more efficient tokenization while reducing disparities across languages. In sequence classification, our method shows improvements over the baselines, yielding higher average F1 scores across benchmarks. Finally, in cross-lingual sentence representation, our approach produces stronger cross-lingual alignment representation than ordinary tokenization methods, making it especially beneficial for cross-lingual training and for incorporating auxiliary languages into encoder-only multilingual models.

We also identify several promising directions for future work. First, an ablation study that investigates the extent to which language identity embeddings contribute to the effectiveness of our approach. Such an analysis would require controlled pretraining experiments, which are computationally expensive and substantial enough to warrant a separate study. Second, extending parallel tokenization to decoder-based models represents another valuable direction, particularly for examining whether its benefits generalize beyond encoder-only multilingual models.

\section*{Limitations}
\label{sec:limitations}




To develop a robust parallel tokenizer, each stage of designing the parallel vocabulary and model input representation required careful consideration. For mapping the English vocabulary in the tokenizer, we employed machine translation (MT). Although bilingual dictionaries can potentially provide more precise word-level mappings due to their predefined vocabularies, our attempt to utilize the PanLex bilingual dictionary (as discussed in Section~\ref{sec:parallel_vocabulary}) revealed substantial limitations in both coverage and data quality, particularly for lower-resource languages. While MT enabled us to align approximately 61\% of tokens, the bilingual dictionary achieved only about 21.5\%. Therefore, we adopted MT for our final approach.

This choice, however, introduces a minor limitation related to the quality of the machine translation itself, as several mapping errors were observed, including invalid outputs such as multiword or malformed phrases. Consequently, there remains potential to improve token alignment through more refined resources or hybrid methods. Furthermore, there is room to enhance the model's input representation, as detailed in Appendix~\ref{sec:proposed_method}, which we plan to investigate further in future work.

Our current study focuses on sequence classification, aligning with our focus on low-resource languages that have limited data. In future research, as more diverse datasets become available, we aim to expand the benchmarking to additional tasks and languages, ensuring that the parallel tokenizer can effectively scale and maintain performance across a broader linguistic spectrum.



\section*{Ethical Consideration}
All datasets used in this work are publicly available and comply with their respective licenses, including Wikipedia, FLORES+, and established benchmarks for sentiment, hate speech, and emotion classification. No personally identifiable or sensitive information was collected or used. Our study aims to improve multilingual representation quality, particularly for low-resource languages. However, since the underlying data may contain cultural or societal biases, we recommend that future work include bias analysis and community engagement when extending models to additional linguistic or cultural contexts.



\bibliography{custom}

@inproceedings{deiseroth-etal-2024-free,
    title = "{T}-{FREE}: Subword Tokenizer-Free Generative {LLM}s via Sparse Representations for Memory-Efficient Embeddings",
    author = {Deiseroth, Bj{\"o}rn  and
      Brack, Manuel  and
      Schramowski, Patrick  and
      Kersting, Kristian  and
      Weinbach, Samuel},
    editor = "Al-Onaizan, Yaser  and
      Bansal, Mohit  and
      Chen, Yun-Nung",
    booktitle = "Proceedings of the 2024 Conference on Empirical Methods in Natural Language Processing",
    month = nov,
    year = "2024",
    address = "Miami, Florida, USA",
    publisher = "Association for Computational Linguistics",
    url = "https://aclanthology.org/2024.emnlp-main.1217/",
    doi = "10.18653/v1/2024.emnlp-main.1217",
    pages = "21829--21851"
}

@inproceedings{kudo-richardson-2018-sentencepiece,
    title = "{S}entence{P}iece: A simple and language independent subword tokenizer and detokenizer for Neural Text Processing",
    author = "Kudo, Taku  and
      Richardson, John",
    editor = "Blanco, Eduardo  and
      Lu, Wei",
    booktitle = "Proceedings of the 2018 Conference on Empirical Methods in Natural Language Processing: System Demonstrations",
    month = nov,
    year = "2018",
    address = "Brussels, Belgium",
    publisher = "Association for Computational Linguistics",
    url = "https://aclanthology.org/D18-2012/",
    doi = "10.18653/v1/D18-2012",
    pages = "66--71"
}

@inproceedings{kudo-2018-subword,
    title = "Subword Regularization: Improving Neural Network Translation Models with Multiple Subword Candidates",
    author = "Kudo, Taku",
    editor = "Gurevych, Iryna  and
      Miyao, Yusuke",
    booktitle = "Proceedings of the 56th Annual Meeting of the Association for Computational Linguistics (Volume 1: Long Papers)",
    month = jul,
    year = "2018",
    address = "Melbourne, Australia",
    publisher = "Association for Computational Linguistics",
    url = "https://aclanthology.org/P18-1007/",
    doi = "10.18653/v1/P18-1007",
    pages = "66--75"
}

@inproceedings{kotosherkala,
  title={Sherkala-Chat: Building a State-of-the-Art LLM for Kazakh in a Moderately Resourced Setting},
  author={Koto, Fajri and Joshi, Rituraj and Mukhituly, Nurdaulet and Wang, Yuxia and Xie, Zhuohan and Pal, Rahul and Orel, Daniil and Mullah, Parvez and Turmakhan, Diana and Goloburda, Maiya and others},
  booktitle={Second Conference on Language Modeling},
  year={2025},
  url={https://openreview.net/forum?id=wRcTCcb0H5}
}

@article{clark-etal-2022-canine,
    title = "Canine: Pre-training an Efficient Tokenization-Free Encoder for Language Representation",
    author = "Clark, Jonathan H.  and
      Garrette, Dan  and
      Turc, Iulia  and
      Wieting, John",
    editor = "Roark, Brian  and
      Nenkova, Ani",
    journal = "Transactions of the Association for Computational Linguistics",
    volume = "10",
    year = "2022",
    address = "Cambridge, MA",
    publisher = "MIT Press",
    url = "https://aclanthology.org/2022.tacl-1.5/",
    doi = "10.1162/tacl_a_00448",
    pages = "73--91"
}

@article{xue-etal-2022-byt5,
    title = "{B}y{T}5: Towards a Token-Free Future with Pre-trained Byte-to-Byte Models",
    author = "Xue, Linting  and
      Barua, Aditya  and
      Constant, Noah  and
      Al-Rfou, Rami  and
      Narang, Sharan  and
      Kale, Mihir  and
      Roberts, Adam  and
      Raffel, Colin",
    editor = "Roark, Brian  and
      Nenkova, Ani",
    journal = "Transactions of the Association for Computational Linguistics",
    volume = "10",
    year = "2022",
    address = "Cambridge, MA",
    publisher = "MIT Press",
    url = "https://aclanthology.org/2022.tacl-1.17/",
    doi = "10.1162/tacl_a_00461",
    pages = "291--306"
}

@inproceedings{cao-2023-best,
    title = "What is the best recipe for character-level encoder-only modelling?",
    author = "Cao, Kris",
    editor = "Rogers, Anna  and
      Boyd-Graber, Jordan  and
      Okazaki, Naoaki",
    booktitle = "Proceedings of the 61st Annual Meeting of the Association for Computational Linguistics (Volume 1: Long Papers)",
    month = jul,
    year = "2023",
    address = "Toronto, Canada",
    publisher = "Association for Computational Linguistics",
    url = "https://aclanthology.org/2023.acl-long.326/",
    doi = "10.18653/v1/2023.acl-long.326",
    pages = "5924--5938"
}

@misc{bai2023qwen,
      title={Qwen Technical Report}, 
      author={Jinze Bai and Shuai Bai and Yunfei Chu and Zeyu Cui and Kai Dang and Xiaodong Deng and Yang Fan and Wenbin Ge and Yu Han and Fei Huang and Binyuan Hui and Luo Ji and Mei Li and Junyang Lin and Runji Lin and Dayiheng Liu and Gao Liu and Chengqiang Lu and Keming Lu and Jianxin Ma and Rui Men and Xingzhang Ren and Xuancheng Ren and Chuanqi Tan and Sinan Tan and Jianhong Tu and Peng Wang and Shijie Wang and Wei Wang and Shengguang Wu and Benfeng Xu and Jin Xu and An Yang and Hao Yang and Jian Yang and Shusheng Yang and Yang Yao and Bowen Yu and Hongyi Yuan and Zheng Yuan and Jianwei Zhang and Xingxuan Zhang and Yichang Zhang and Zhenru Zhang and Chang Zhou and Jingren Zhou and Xiaohuan Zhou and Tianhang Zhu},
      year={2023},
      eprint={2309.16609},
      archivePrefix={arXiv},
      primaryClass={cs.CL},
      url={https://arxiv.org/abs/2309.16609}, 
}

@misc{achiam2023gpt,
      title={GPT-4 Technical Report}, 
      author={OpenAI and Josh Achiam and Steven Adler and Sandhini Agarwal and Lama Ahmad and Ilge Akkaya and Florencia Leoni Aleman and Diogo Almeida and Janko Altenschmidt and Sam Altman and Shyamal Anadkat and Red Avila and Igor Babuschkin and Suchir Balaji and Valerie Balcom and Paul Baltescu and Haiming Bao and Mohammad Bavarian and Jeff Belgum and Irwan Bello and Jake Berdine and Gabriel Bernadett-Shapiro and Christopher Berner and Lenny Bogdonoff and Oleg Boiko and Madelaine Boyd and Anna-Luisa Brakman and Greg Brockman and Tim Brooks and Miles Brundage and Kevin Button and Trevor Cai and Rosie Campbell and Andrew Cann and Brittany Carey and Chelsea Carlson and Rory Carmichael and Brooke Chan and Che Chang and Fotis Chantzis and Derek Chen and Sully Chen and Ruby Chen and Jason Chen and Mark Chen and Ben Chess and Chester Cho and Casey Chu and Hyung Won Chung and Dave Cummings and Jeremiah Currier and Yunxing Dai and Cory Decareaux and Thomas Degry and Noah Deutsch and Damien Deville and Arka Dhar and David Dohan and Steve Dowling and Sheila Dunning and Adrien Ecoffet and Atty Eleti and Tyna Eloundou and David Farhi and Liam Fedus and Niko Felix and Simón Posada Fishman and Juston Forte and Isabella Fulford and Leo Gao and Elie Georges and Christian Gibson and Vik Goel and Tarun Gogineni and Gabriel Goh and Rapha Gontijo-Lopes and Jonathan Gordon and Morgan Grafstein and Scott Gray and Ryan Greene and Joshua Gross and Shixiang Shane Gu and Yufei Guo and Chris Hallacy and Jesse Han and Jeff Harris and Yuchen He and Mike Heaton and Johannes Heidecke and Chris Hesse and Alan Hickey and Wade Hickey and Peter Hoeschele and Brandon Houghton and Kenny Hsu and Shengli Hu and Xin Hu and Joost Huizinga and Shantanu Jain and Shawn Jain and Joanne Jang and Angela Jiang and Roger Jiang and Haozhun Jin and Denny Jin and Shino Jomoto and Billie Jonn and Heewoo Jun and Tomer Kaftan and Łukasz Kaiser and Ali Kamali and Ingmar Kanitscheider and Nitish Shirish Keskar and Tabarak Khan and Logan Kilpatrick and Jong Wook Kim and Christina Kim and Yongjik Kim and Jan Hendrik Kirchner and Jamie Kiros and Matt Knight and Daniel Kokotajlo and Łukasz Kondraciuk and Andrew Kondrich and Aris Konstantinidis and Kyle Kosic and Gretchen Krueger and Vishal Kuo and Michael Lampe and Ikai Lan and Teddy Lee and Jan Leike and Jade Leung and Daniel Levy and Chak Ming Li and Rachel Lim and Molly Lin and Stephanie Lin and Mateusz Litwin and Theresa Lopez and Ryan Lowe and Patricia Lue and Anna Makanju and Kim Malfacini and Sam Manning and Todor Markov and Yaniv Markovski and Bianca Martin and Katie Mayer and Andrew Mayne and Bob McGrew and Scott Mayer McKinney and Christine McLeavey and Paul McMillan and Jake McNeil and David Medina and Aalok Mehta and Jacob Menick and Luke Metz and Andrey Mishchenko and Pamela Mishkin and Vinnie Monaco and Evan Morikawa and Daniel Mossing and Tong Mu and Mira Murati and Oleg Murk and David Mély and Ashvin Nair and Reiichiro Nakano and Rajeev Nayak and Arvind Neelakantan and Richard Ngo and Hyeonwoo Noh and Long Ouyang and Cullen O'Keefe and Jakub Pachocki and Alex Paino and Joe Palermo and Ashley Pantuliano and Giambattista Parascandolo and Joel Parish and Emy Parparita and Alex Passos and Mikhail Pavlov and Andrew Peng and Adam Perelman and Filipe de Avila Belbute Peres and Michael Petrov and Henrique Ponde de Oliveira Pinto and Michael and Pokorny and Michelle Pokrass and Vitchyr H. Pong and Tolly Powell and Alethea Power and Boris Power and Elizabeth Proehl and Raul Puri and Alec Radford and Jack Rae and Aditya Ramesh and Cameron Raymond and Francis Real and Kendra Rimbach and Carl Ross and Bob Rotsted and Henri Roussez and Nick Ryder and Mario Saltarelli and Ted Sanders and Shibani Santurkar and Girish Sastry and Heather Schmidt and David Schnurr and John Schulman and Daniel Selsam and Kyla Sheppard and Toki Sherbakov and Jessica Shieh and Sarah Shoker and Pranav Shyam and Szymon Sidor and Eric Sigler and Maddie Simens and Jordan Sitkin and Katarina Slama and Ian Sohl and Benjamin Sokolowsky and Yang Song and Natalie Staudacher and Felipe Petroski Such and Natalie Summers and Ilya Sutskever and Jie Tang and Nikolas Tezak and Madeleine B. Thompson and Phil Tillet and Amin Tootoonchian and Elizabeth Tseng and Preston Tuggle and Nick Turley and Jerry Tworek and Juan Felipe Cerón Uribe and Andrea Vallone and Arun Vijayvergiya and Chelsea Voss and Carroll Wainwright and Justin Jay Wang and Alvin Wang and Ben Wang and Jonathan Ward and Jason Wei and CJ Weinmann and Akila Welihinda and Peter Welinder and Jiayi Weng and Lilian Weng and Matt Wiethoff and Dave Willner and Clemens Winter and Samuel Wolrich and Hannah Wong and Lauren Workman and Sherwin Wu and Jeff Wu and Michael Wu and Kai Xiao and Tao Xu and Sarah Yoo and Kevin Yu and Qiming Yuan and Wojciech Zaremba and Rowan Zellers and Chong Zhang and Marvin Zhang and Shengjia Zhao and Tianhao Zheng and Juntang Zhuang and William Zhuk and Barret Zoph},
      year={2024},
      eprint={2303.08774},
      archivePrefix={arXiv},
      primaryClass={cs.CL},
      url={https://arxiv.org/abs/2303.08774}, 
}

@misc{dubey2024llama,
      title={The Llama 3 Herd of Models}, 
      author={Aaron Grattafiori and Abhimanyu Dubey and Abhinav Jauhri and Abhinav Pandey and Abhishek Kadian and Ahmad Al-Dahle and Aiesha Letman and Akhil Mathur and Alan Schelten and Alex Vaughan and Amy Yang and Angela Fan and Anirudh Goyal and Anthony Hartshorn and Aobo Yang and Archi Mitra and Archie Sravankumar and Artem Korenev and Arthur Hinsvark and Arun Rao and Aston Zhang and Aurelien Rodriguez and Austen Gregerson and Ava Spataru and Baptiste Roziere and Bethany Biron and Binh Tang and Bobbie Chern and Charlotte Caucheteux and Chaya Nayak and Chloe Bi and Chris Marra and Chris McConnell and Christian Keller and Christophe Touret and Chunyang Wu and Corinne Wong and Cristian Canton Ferrer and Cyrus Nikolaidis and Damien Allonsius and Daniel Song and Danielle Pintz and Danny Livshits and Danny Wyatt and David Esiobu and Dhruv Choudhary and Dhruv Mahajan and Diego Garcia-Olano and Diego Perino and Dieuwke Hupkes and Egor Lakomkin and Ehab AlBadawy and Elina Lobanova and Emily Dinan and Eric Michael Smith and Filip Radenovic and Francisco Guzmán and Frank Zhang and Gabriel Synnaeve and Gabrielle Lee and Georgia Lewis Anderson and Govind Thattai and Graeme Nail and Gregoire Mialon and Guan Pang and Guillem Cucurell and Hailey Nguyen and Hannah Korevaar and Hu Xu and Hugo Touvron and Iliyan Zarov and Imanol Arrieta Ibarra and Isabel Kloumann and Ishan Misra and Ivan Evtimov and Jack Zhang and Jade Copet and Jaewon Lee and Jan Geffert and Jana Vranes and Jason Park and Jay Mahadeokar and Jeet Shah and Jelmer van der Linde and Jennifer Billock and Jenny Hong and Jenya Lee and Jeremy Fu and Jianfeng Chi and Jianyu Huang and Jiawen Liu and Jie Wang and Jiecao Yu and Joanna Bitton and Joe Spisak and Jongsoo Park and Joseph Rocca and Joshua Johnstun and Joshua Saxe and Junteng Jia and Kalyan Vasuden Alwala and Karthik Prasad and Kartikeya Upasani and Kate Plawiak and Ke Li and Kenneth Heafield and Kevin Stone and Khalid El-Arini and Krithika Iyer and Kshitiz Malik and Kuenley Chiu and Kunal Bhalla and Kushal Lakhotia and Lauren Rantala-Yeary and Laurens van der Maaten and Lawrence Chen and Liang Tan and Liz Jenkins and Louis Martin and Lovish Madaan and Lubo Malo and Lukas Blecher and Lukas Landzaat and Luke de Oliveira and Madeline Muzzi and Mahesh Pasupuleti and Mannat Singh and Manohar Paluri and Marcin Kardas and Maria Tsimpoukelli and Mathew Oldham and Mathieu Rita and Maya Pavlova and Melanie Kambadur and Mike Lewis and Min Si and Mitesh Kumar Singh and Mona Hassan and Naman Goyal and Narjes Torabi and Nikolay Bashlykov and Nikolay Bogoychev and Niladri Chatterji and Ning Zhang and Olivier Duchenne and Onur Çelebi and Patrick Alrassy and Pengchuan Zhang and Pengwei Li and Petar Vasic and Peter Weng and Prajjwal Bhargava and Pratik Dubal and Praveen Krishnan and Punit Singh Koura and Puxin Xu and Qing He and Qingxiao Dong and Ragavan Srinivasan and Raj Ganapathy and Ramon Calderer and Ricardo Silveira Cabral and Robert Stojnic and Roberta Raileanu and Rohan Maheswari and Rohit Girdhar and Rohit Patel and Romain Sauvestre and Ronnie Polidoro and Roshan Sumbaly and Ross Taylor and Ruan Silva and Rui Hou and Rui Wang and Saghar Hosseini and Sahana Chennabasappa and Sanjay Singh and Sean Bell and Seohyun Sonia Kim and Sergey Edunov and Shaoliang Nie and Sharan Narang and Sharath Raparthy and Sheng Shen and Shengye Wan and Shruti Bhosale and Shun Zhang and Simon Vandenhende and Soumya Batra and Spencer Whitman and Sten Sootla and Stephane Collot and Suchin Gururangan and Sydney Borodinsky and Tamar Herman and Tara Fowler and Tarek Sheasha and Thomas Georgiou and Thomas Scialom and Tobias Speckbacher and Todor Mihaylov and Tong Xiao and Ujjwal Karn and Vedanuj Goswami and Vibhor Gupta and Vignesh Ramanathan and Viktor Kerkez and Vincent Gonguet and Virginie Do and Vish Vogeti and Vítor Albiero and Vladan Petrovic and Weiwei Chu and Wenhan Xiong and Wenyin Fu and Whitney Meers and Xavier Martinet and Xiaodong Wang and Xiaofang Wang and Xiaoqing Ellen Tan and Xide Xia and Xinfeng Xie and Xuchao Jia and Xuewei Wang and Yaelle Goldschlag and Yashesh Gaur and Yasmine Babaei and Yi Wen and Yiwen Song and Yuchen Zhang and Yue Li and Yuning Mao and Zacharie Delpierre Coudert and Zheng Yan and Zhengxing Chen and Zoe Papakipos and Aaditya Singh and Aayushi Srivastava and Abha Jain and Adam Kelsey and Adam Shajnfeld and Adithya Gangidi and Adolfo Victoria and Ahuva Goldstand and Ajay Menon and Ajay Sharma and Alex Boesenberg and Alexei Baevski and Allie Feinstein and Amanda Kallet and Amit Sangani and Amos Teo and Anam Yunus and Andrei Lupu and Andres Alvarado and Andrew Caples and Andrew Gu and Andrew Ho and Andrew Poulton and Andrew Ryan and Ankit Ramchandani and Annie Dong and Annie Franco and Anuj Goyal and Aparajita Saraf and Arkabandhu Chowdhury and Ashley Gabriel and Ashwin Bharambe and Assaf Eisenman and Azadeh Yazdan and Beau James and Ben Maurer and Benjamin Leonhardi and Bernie Huang and Beth Loyd and Beto De Paola and Bhargavi Paranjape and Bing Liu and Bo Wu and Boyu Ni and Braden Hancock and Bram Wasti and Brandon Spence and Brani Stojkovic and Brian Gamido and Britt Montalvo and Carl Parker and Carly Burton and Catalina Mejia and Ce Liu and Changhan Wang and Changkyu Kim and Chao Zhou and Chester Hu and Ching-Hsiang Chu and Chris Cai and Chris Tindal and Christoph Feichtenhofer and Cynthia Gao and Damon Civin and Dana Beaty and Daniel Kreymer and Daniel Li and David Adkins and David Xu and Davide Testuggine and Delia David and Devi Parikh and Diana Liskovich and Didem Foss and Dingkang Wang and Duc Le and Dustin Holland and Edward Dowling and Eissa Jamil and Elaine Montgomery and Eleonora Presani and Emily Hahn and Emily Wood and Eric-Tuan Le and Erik Brinkman and Esteban Arcaute and Evan Dunbar and Evan Smothers and Fei Sun and Felix Kreuk and Feng Tian and Filippos Kokkinos and Firat Ozgenel and Francesco Caggioni and Frank Kanayet and Frank Seide and Gabriela Medina Florez and Gabriella Schwarz and Gada Badeer and Georgia Swee and Gil Halpern and Grant Herman and Grigory Sizov and Guangyi and Zhang and Guna Lakshminarayanan and Hakan Inan and Hamid Shojanazeri and Han Zou and Hannah Wang and Hanwen Zha and Haroun Habeeb and Harrison Rudolph and Helen Suk and Henry Aspegren and Hunter Goldman and Hongyuan Zhan and Ibrahim Damlaj and Igor Molybog and Igor Tufanov and Ilias Leontiadis and Irina-Elena Veliche and Itai Gat and Jake Weissman and James Geboski and James Kohli and Janice Lam and Japhet Asher and Jean-Baptiste Gaya and Jeff Marcus and Jeff Tang and Jennifer Chan and Jenny Zhen and Jeremy Reizenstein and Jeremy Teboul and Jessica Zhong and Jian Jin and Jingyi Yang and Joe Cummings and Jon Carvill and Jon Shepard and Jonathan McPhie and Jonathan Torres and Josh Ginsburg and Junjie Wang and Kai Wu and Kam Hou U and Karan Saxena and Kartikay Khandelwal and Katayoun Zand and Kathy Matosich and Kaushik Veeraraghavan and Kelly Michelena and Keqian Li and Kiran Jagadeesh and Kun Huang and Kunal Chawla and Kyle Huang and Lailin Chen and Lakshya Garg and Lavender A and Leandro Silva and Lee Bell and Lei Zhang and Liangpeng Guo and Licheng Yu and Liron Moshkovich and Luca Wehrstedt and Madian Khabsa and Manav Avalani and Manish Bhatt and Martynas Mankus and Matan Hasson and Matthew Lennie and Matthias Reso and Maxim Groshev and Maxim Naumov and Maya Lathi and Meghan Keneally and Miao Liu and Michael L. Seltzer and Michal Valko and Michelle Restrepo and Mihir Patel and Mik Vyatskov and Mikayel Samvelyan and Mike Clark and Mike Macey and Mike Wang and Miquel Jubert Hermoso and Mo Metanat and Mohammad Rastegari and Munish Bansal and Nandhini Santhanam and Natascha Parks and Natasha White and Navyata Bawa and Nayan Singhal and Nick Egebo and Nicolas Usunier and Nikhil Mehta and Nikolay Pavlovich Laptev and Ning Dong and Norman Cheng and Oleg Chernoguz and Olivia Hart and Omkar Salpekar and Ozlem Kalinli and Parkin Kent and Parth Parekh and Paul Saab and Pavan Balaji and Pedro Rittner and Philip Bontrager and Pierre Roux and Piotr Dollar and Polina Zvyagina and Prashant Ratanchandani and Pritish Yuvraj and Qian Liang and Rachad Alao and Rachel Rodriguez and Rafi Ayub and Raghotham Murthy and Raghu Nayani and Rahul Mitra and Rangaprabhu Parthasarathy and Raymond Li and Rebekkah Hogan and Robin Battey and Rocky Wang and Russ Howes and Ruty Rinott and Sachin Mehta and Sachin Siby and Sai Jayesh Bondu and Samyak Datta and Sara Chugh and Sara Hunt and Sargun Dhillon and Sasha Sidorov and Satadru Pan and Saurabh Mahajan and Saurabh Verma and Seiji Yamamoto and Sharadh Ramaswamy and Shaun Lindsay and Shaun Lindsay and Sheng Feng and Shenghao Lin and Shengxin Cindy Zha and Shishir Patil and Shiva Shankar and Shuqiang Zhang and Shuqiang Zhang and Sinong Wang and Sneha Agarwal and Soji Sajuyigbe and Soumith Chintala and Stephanie Max and Stephen Chen and Steve Kehoe and Steve Satterfield and Sudarshan Govindaprasad and Sumit Gupta and Summer Deng and Sungmin Cho and Sunny Virk and Suraj Subramanian and Sy Choudhury and Sydney Goldman and Tal Remez and Tamar Glaser and Tamara Best and Thilo Koehler and Thomas Robinson and Tianhe Li and Tianjun Zhang and Tim Matthews and Timothy Chou and Tzook Shaked and Varun Vontimitta and Victoria Ajayi and Victoria Montanez and Vijai Mohan and Vinay Satish Kumar and Vishal Mangla and Vlad Ionescu and Vlad Poenaru and Vlad Tiberiu Mihailescu and Vladimir Ivanov and Wei Li and Wenchen Wang and Wenwen Jiang and Wes Bouaziz and Will Constable and Xiaocheng Tang and Xiaojian Wu and Xiaolan Wang and Xilun Wu and Xinbo Gao and Yaniv Kleinman and Yanjun Chen and Ye Hu and Ye Jia and Ye Qi and Yenda Li and Yilin Zhang and Ying Zhang and Yossi Adi and Youngjin Nam and Yu and Wang and Yu Zhao and Yuchen Hao and Yundi Qian and Yunlu Li and Yuzi He and Zach Rait and Zachary DeVito and Zef Rosnbrick and Zhaoduo Wen and Zhenyu Yang and Zhiwei Zhao and Zhiyu Ma},
      year={2024},
      eprint={2407.21783},
      archivePrefix={arXiv},
      primaryClass={cs.AI},
      url={https://arxiv.org/abs/2407.21783}, 
}

@inproceedings{rust-etal-2021-good,
    title = "How Good is Your Tokenizer? On the Monolingual Performance of Multilingual Language Models",
    author = "Rust, Phillip  and
      Pfeiffer, Jonas  and
      Vuli{\'c}, Ivan  and
      Ruder, Sebastian  and
      Gurevych, Iryna",
    editor = "Zong, Chengqing  and
      Xia, Fei  and
      Li, Wenjie  and
      Navigli, Roberto",
    booktitle = "Proceedings of the 59th Annual Meeting of the Association for Computational Linguistics and the 11th International Joint Conference on Natural Language Processing (Volume 1: Long Papers)",
    month = aug,
    year = "2021",
    address = "Online",
    publisher = "Association for Computational Linguistics",
    url = "https://aclanthology.org/2021.acl-long.243/",
    doi = "10.18653/v1/2021.acl-long.243",
    pages = "3118--3135"
}

@inproceedings{xue-etal-2021-mt5,
    title = "m{T}5: A Massively Multilingual Pre-trained Text-to-Text Transformer",
    author = "Xue, Linting  and
      Constant, Noah  and
      Roberts, Adam  and
      Kale, Mihir  and
      Al-Rfou, Rami  and
      Siddhant, Aditya  and
      Barua, Aditya  and
      Raffel, Colin",
    editor = "Toutanova, Kristina  and
      Rumshisky, Anna  and
      Zettlemoyer, Luke  and
      Hakkani-Tur, Dilek  and
      Beltagy, Iz  and
      Bethard, Steven  and
      Cotterell, Ryan  and
      Chakraborty, Tanmoy  and
      Zhou, Yichao",
    booktitle = "Proceedings of the 2021 Conference of the North American Chapter of the Association for Computational Linguistics: Human Language Technologies",
    month = jun,
    year = "2021",
    address = "Online",
    publisher = "Association for Computational Linguistics",
    url = "https://aclanthology.org/2021.naacl-main.41/",
    doi = "10.18653/v1/2021.naacl-main.41",
    pages = "483--498"
}

@inproceedings{conneau-etal-2020-unsupervised,
    title = "Unsupervised Cross-lingual Representation Learning at Scale",
    author = "Conneau, Alexis  and
      Khandelwal, Kartikay  and
      Goyal, Naman  and
      Chaudhary, Vishrav  and
      Wenzek, Guillaume  and
      Guzm{\'a}n, Francisco  and
      Grave, Edouard  and
      Ott, Myle  and
      Zettlemoyer, Luke  and
      Stoyanov, Veselin",
    editor = "Jurafsky, Dan  and
      Chai, Joyce  and
      Schluter, Natalie  and
      Tetreault, Joel",
    booktitle = "Proceedings of the 58th Annual Meeting of the Association for Computational Linguistics",
    month = jul,
    year = "2020",
    address = "Online",
    publisher = "Association for Computational Linguistics",
    url = "https://aclanthology.org/2020.acl-main.747/",
    doi = "10.18653/v1/2020.acl-main.747",
    pages = "8440--8451"
}

@inproceedings{devlin-etal-2019-bert,
    title = "{BERT}: Pre-training of Deep Bidirectional Transformers for Language Understanding",
    author = "Devlin, Jacob  and
      Chang, Ming-Wei  and
      Lee, Kenton  and
      Toutanova, Kristina",
    editor = "Burstein, Jill  and
      Doran, Christy  and
      Solorio, Thamar",
    booktitle = "Proceedings of the 2019 Conference of the North {A}merican Chapter of the Association for Computational Linguistics: Human Language Technologies, Volume 1 (Long and Short Papers)",
    month = jun,
    year = "2019",
    address = "Minneapolis, Minnesota",
    publisher = "Association for Computational Linguistics",
    url = "https://aclanthology.org/N19-1423/",
    doi = "10.18653/v1/N19-1423",
    pages = "4171--4186"
}

@inproceedings{sennrich-etal-2016-neural,
    title = "Neural Machine Translation of Rare Words with Subword Units",
    author = "Sennrich, Rico  and
      Haddow, Barry  and
      Birch, Alexandra",
    editor = "Erk, Katrin  and
      Smith, Noah A.",
    booktitle = "Proceedings of the 54th Annual Meeting of the Association for Computational Linguistics (Volume 1: Long Papers)",
    month = aug,
    year = "2016",
    address = "Berlin, Germany",
    publisher = "Association for Computational Linguistics",
    url = "https://aclanthology.org/P16-1162/",
    doi = "10.18653/v1/P16-1162",
    pages = "1715--1725"
}

@misc{costa2022no,
      title={No Language Left Behind: Scaling Human-Centered Machine Translation}, 
      author={NLLB Team and Marta R. Costa-jussa and James Cross and Onur Celebi and Maha Elbayad and Kenneth Heafield and Kevin Heffernan and Elahe Kalbassi and Janice Lam and Daniel Licht and Jean Maillard and Anna Sun and Skyler Wang and Guillaume Wenzek and Al Youngblood and Bapi Akula and Loic Barrault and Gabriel Mejia Gonzalez and Prangthip Hansanti and John Hoffman and Semarley Jarrett and Kaushik Ram Sadagopan and Dirk Rowe and Shannon Spruit and Chau Tran and Pierre Andrews and Necip Fazil Ayan and Shruti Bhosale and Sergey Edunov and Angela Fan and Cynthia Gao and Vedanuj Goswami and Francisco Guzmán and Philipp Koehn and Alexandre Mourachko and Christophe Ropers and Safiyyah Saleem and Holger Schwenk and Jeff Wang},
      year={2022},
      eprint={2207.04672},
      archivePrefix={arXiv},
      primaryClass={cs.CL},
      url={https://arxiv.org/abs/2207.04672}, 
}

@inproceedings{winata-etal-2023-nusax,
    title = "{N}usa{X}: Multilingual Parallel Sentiment Dataset for 10 {I}ndonesian Local Languages",
    author = "Winata, Genta Indra  and
      Aji, Alham Fikri  and
      Cahyawijaya, Samuel  and
      Mahendra, Rahmad  and
      Koto, Fajri  and
      Romadhony, Ade  and
      Kurniawan, Kemal  and
      Moeljadi, David  and
      Prasojo, Radityo Eko  and
      Fung, Pascale  and
      Baldwin, Timothy  and
      Lau, Jey Han  and
      Sennrich, Rico  and
      Ruder, Sebastian",
    editor = "Vlachos, Andreas  and
      Augenstein, Isabelle",
    booktitle = "Proceedings of the 17th Conference of the European Chapter of the Association for Computational Linguistics",
    month = may,
    year = "2023",
    address = "Dubrovnik, Croatia",
    publisher = "Association for Computational Linguistics",
    url = "https://aclanthology.org/2023.eacl-main.57/",
    doi = "10.18653/v1/2023.eacl-main.57",
    pages = "815--834"
}

@inproceedings{muhammad-etal-2025-afrihate,
    title = "{A}fri{H}ate: A Multilingual Collection of Hate Speech and Abusive Language Datasets for {A}frican Languages",
    author = {Muhammad, Shamsuddeen Hassan  and
      Abdulmumin, Idris  and
      Ayele, Abinew Ali  and
      Adelani, David Ifeoluwa  and
      Ahmad, Ibrahim Said  and
      Aliyu, Saminu Mohammad  and
      R{\"o}ttger, Paul  and
      Oppong, Abigail  and
      Bukula, Andiswa  and
      Chukwuneke, Chiamaka Ijeoma  and
      Jibril, Ebrahim Chekol  and
      Ismail, Elyas Abdi  and
      Alemneh, Esubalew  and
      Gebremichael, Hagos Tesfahun  and
      Aliyu, Lukman Jibril  and
      Beloucif, Meriem  and
      Hourrane, Oumaima  and
      Mabuya, Rooweither  and
      Osei, Salomey  and
      Rutunda, Samuel  and
      Belay, Tadesse Destaw  and
      Guge, Tadesse Kebede  and
      Asfaw, Tesfa Tegegne  and
      Wanzare, Lilian Diana Awuor  and
      Onyango, Nelson Odhiambo  and
      Yimam, Seid Muhie  and
      Ousidhoum, Nedjma},
    editor = "Chiruzzo, Luis  and
      Ritter, Alan  and
      Wang, Lu",
    booktitle = "Proceedings of the 2025 Conference of the Nations of the Americas Chapter of the Association for Computational Linguistics: Human Language Technologies (Volume 1: Long Papers)",
    month = apr,
    year = "2025",
    address = "Albuquerque, New Mexico",
    publisher = "Association for Computational Linguistics",
    url = "https://aclanthology.org/2025.naacl-long.92/",
    doi = "10.18653/v1/2025.naacl-long.92",
    pages = "1854--1871",
    ISBN = "979-8-89176-189-6"
}

@inproceedings{belay-etal-2025-ethioemo,
    title = "Evaluating the Capabilities of Large Language Models for Multi-label Emotion Understanding",
    author = "Belay, Tadesse Destaw  and
      Azime, Israel Abebe  and
      Ayele, Abinew Ali  and
      Sidorov, Grigori  and
      Klakow, Dietrich  and
      Slusallek, Philip  and
      Kolesnikova, Olga  and
      Yimam, Seid Muhie",
    editor = "Rambow, Owen  and
      Wanner, Leo  and
      Apidianaki, Marianna  and
      Al-Khalifa, Hend  and
      Eugenio, Barbara Di  and
      Schockaert, Steven",
    booktitle = "Proceedings of the 31st International Conference on Computational Linguistics",
    month = jan,
    year = "2025",
    address = "Abu Dhabi, UAE",
    publisher = "Association for Computational Linguistics",
    url = "https://aclanthology.org/2025.coling-main.237/",
    pages = "3523--3540"
}

@inproceedings{muhammad-etal-2025-brighter,
    title = "{BRIGHTER}: {BRI}dging the Gap in Human-Annotated Textual Emotion Recognition Datasets for 28 Languages",
    author = "Muhammad, Shamsuddeen Hassan  and
      Ousidhoum, Nedjma  and
      Abdulmumin, Idris  and
      Wahle, Jan Philip  and
      Ruas, Terry  and
      Beloucif, Meriem  and
      de Kock, Christine  and
      Surange, Nirmal  and
      Teodorescu, Daniela  and
      Ahmad, Ibrahim Said  and
      Adelani, David Ifeoluwa  and
      Aji, Alham Fikri  and
      Ali, Felermino D. M. A.  and
      Alimova, Ilseyar  and
      Araujo, Vladimir  and
      Babakov, Nikolay  and
      Baes, Naomi  and
      Bucur, Ana-Maria  and
      Bukula, Andiswa  and
      Cao, Guanqun  and
      Tufi{\~n}o, Rodrigo  and
      Chevi, Rendi  and
      Chukwuneke, Chiamaka Ijeoma  and
      Ciobotaru, Alexandra  and
      Dementieva, Daryna  and
      Gadanya, Murja Sani  and
      Geislinger, Robert  and
      Gipp, Bela  and
      Hourrane, Oumaima  and
      Ignat, Oana  and
      Lawan, Falalu Ibrahim  and
      Mabuya, Rooweither  and
      Mahendra, Rahmad  and
      Marivate, Vukosi  and
      Panchenko, Alexander  and
      Piper, Andrew  and
      Ferreira, Charles Henrique Porto  and
      Protasov, Vitaly  and
      Rutunda, Samuel  and
      Shrivastava, Manish  and
      Udrea, Aura Cristina  and
      Wanzare, Lilian Diana Awuor  and
      Wu, Sophie  and
      Wunderlich, Florian Valentin  and
      Zhafran, Hanif Muhammad  and
      Zhang, Tianhui  and
      Zhou, Yi  and
      Mohammad, Saif M.",
    editor = "Che, Wanxiang  and
      Nabende, Joyce  and
      Shutova, Ekaterina  and
      Pilehvar, Mohammad Taher",
    booktitle = "Proceedings of the 63rd Annual Meeting of the Association for Computational Linguistics (Volume 1: Long Papers)",
    month = jul,
    year = "2025",
    address = "Vienna, Austria",
    publisher = "Association for Computational Linguistics",
    url = "https://aclanthology.org/2025.acl-long.436/",
    doi = "10.18653/v1/2025.acl-long.436",
    pages = "8895--8916",
    ISBN = "979-8-89176-251-0"
}

@inproceedings{petrov-etal-2023-language,
 author = {Petrov, Aleksandar and La Malfa, Emanuele and Torr, Philip and Bibi, Adel},
 booktitle = {Advances in Neural Information Processing Systems},
 editor = {A. Oh and T. Naumann and A. Globerson and K. Saenko and M. Hardt and S. Levine},
 pages = {36963--36990},
 publisher = {Curran Associates, Inc.},
 title = {Language Model Tokenizers Introduce Unfairness Between Languages},
 url = {https://proceedings.neurips.cc/paper_files/paper/2023/file/74bb24dca8334adce292883b4b651eda-Paper-Conference.pdf},
 volume = {36},
 year = {2023}
}

@inproceedings{huang-etal-2025-modular,
    title = "Modular Sentence Encoders: Separating Language Specialization from Cross-Lingual Alignment",
    author = "Huang, Yongxin  and
      Wang, Kexin  and
      Glava{\v{s}}, Goran  and
      Gurevych, Iryna",
    editor = "Che, Wanxiang  and
      Nabende, Joyce  and
      Shutova, Ekaterina  and
      Pilehvar, Mohammad Taher",
    booktitle = "Proceedings of the 63rd Annual Meeting of the Association for Computational Linguistics (Volume 1: Long Papers)",
    month = jul,
    year = "2025",
    address = "Vienna, Austria",
    publisher = "Association for Computational Linguistics",
    url = "https://aclanthology.org/2025.acl-long.108/",
    doi = "10.18653/v1/2025.acl-long.108",
    pages = "2167--2187",
    ISBN = "979-8-89176-251-0"
}

@inproceedings{artetxe-schwenk-2019-margin,
    title = "Margin-based Parallel Corpus Mining with Multilingual Sentence Embeddings",
    author = "Artetxe, Mikel  and
      Schwenk, Holger",
    editor = "Korhonen, Anna  and
      Traum, David  and
      M{\`a}rquez, Llu{\'i}s",
    booktitle = "Proceedings of the 57th Annual Meeting of the Association for Computational Linguistics",
    month = jul,
    year = "2019",
    address = "Florence, Italy",
    publisher = "Association for Computational Linguistics",
    url = "https://aclanthology.org/P19-1309/",
    doi = "10.18653/v1/P19-1309",
    pages = "3197--3203"
}

@inproceedings{koto-etal-2021-indobertweet,
    title = "{I}ndo{BERT}weet: A Pretrained Language Model for {I}ndonesian {T}witter with Effective Domain-Specific Vocabulary Initialization",
    author = "Koto, Fajri  and
      Lau, Jey Han  and
      Baldwin, Timothy",
    editor = "Moens, Marie-Francine  and
      Huang, Xuanjing  and
      Specia, Lucia  and
      Yih, Scott Wen-tau",
    booktitle = "Proceedings of the 2021 Conference on Empirical Methods in Natural Language Processing",
    month = nov,
    year = "2021",
    address = "Online and Punta Cana, Dominican Republic",
    publisher = "Association for Computational Linguistics",
    url = "https://aclanthology.org/2021.emnlp-main.833/",
    doi = "10.18653/v1/2021.emnlp-main.833",
    pages = "10660--10668"
}

@inproceedings{conneau-etal-2020-emerging,
    title = "Emerging Cross-lingual Structure in Pretrained Language Models",
    author = "Conneau, Alexis  and
      Wu, Shijie  and
      Li, Haoran  and
      Zettlemoyer, Luke  and
      Stoyanov, Veselin",
    editor = "Jurafsky, Dan  and
      Chai, Joyce  and
      Schluter, Natalie  and
      Tetreault, Joel",
    booktitle = "Proceedings of the 58th Annual Meeting of the Association for Computational Linguistics",
    month = jul,
    year = "2020",
    address = "Online",
    publisher = "Association for Computational Linguistics",
    url = "https://aclanthology.org/2020.acl-main.536/",
    doi = "10.18653/v1/2020.acl-main.536",
    pages = "6022--6034"
}

@inproceedings{rajaraman-etal-2024-markov,
 author = {Rajaraman, Nived and Jiao, Jiantao and Ramchandran, Kannan},
 booktitle = {Advances in Neural Information Processing Systems},
 editor = {A. Globerson and L. Mackey and D. Belgrave and A. Fan and U. Paquet and J. Tomczak and C. Zhang},
 pages = {62503--62556},
 publisher = {Curran Associates, Inc.},
 title = {An Analysis of Tokenization: Transformers under Markov Data},
 url = {https://proceedings.neurips.cc/paper_files/paper/2024/file/724afcaae4ae92a9220a077ffe80088d-Paper-Conference.pdf},
 volume = {37},
 year = {2024}
}

@inproceedings{limisiewicz-etal-2023-tokenization,
    title = "Tokenization Impacts Multilingual Language Modeling: Assessing Vocabulary Allocation and Overlap Across Languages",
    author = "Limisiewicz, Tomasz  and
      Balhar, Ji{\v{r}}{\'i}  and
      Mare{\v{c}}ek, David",
    editor = "Rogers, Anna  and
      Boyd-Graber, Jordan  and
      Okazaki, Naoaki",
    booktitle = "Findings of the Association for Computational Linguistics: ACL 2023",
    month = jul,
    year = "2023",
    address = "Toronto, Canada",
    publisher = "Association for Computational Linguistics",
    url = "https://aclanthology.org/2023.findings-acl.350/",
    doi = "10.18653/v1/2023.findings-acl.350",
    pages = "5661--5681"
}

@inproceedings{zheng-etal-2021-allocating,
    title = "Allocating Large Vocabulary Capacity for Cross-Lingual Language Model Pre-Training",
    author = "Zheng, Bo  and
      Dong, Li  and
      Huang, Shaohan  and
      Singhal, Saksham  and
      Che, Wanxiang  and
      Liu, Ting  and
      Song, Xia  and
      Wei, Furu",
    editor = "Moens, Marie-Francine  and
      Huang, Xuanjing  and
      Specia, Lucia  and
      Yih, Scott Wen-tau",
    booktitle = "Proceedings of the 2021 Conference on Empirical Methods in Natural Language Processing",
    month = nov,
    year = "2021",
    address = "Online and Punta Cana, Dominican Republic",
    publisher = "Association for Computational Linguistics",
    url = "https://aclanthology.org/2021.emnlp-main.257/",
    doi = "10.18653/v1/2021.emnlp-main.257",
    pages = "3203--3215"
}

@inproceedings{chung-etal-2020-improving,
    title = "Improving Multilingual Models with Language-Clustered Vocabularies",
    author = "Chung, Hyung Won  and
      Garrette, Dan  and
      Tan, Kiat Chuan  and
      Riesa, Jason",
    editor = "Webber, Bonnie  and
      Cohn, Trevor  and
      He, Yulan  and
      Liu, Yang",
    booktitle = "Proceedings of the 2020 Conference on Empirical Methods in Natural Language Processing (EMNLP)",
    month = nov,
    year = "2020",
    address = "Online",
    publisher = "Association for Computational Linguistics",
    url = "https://aclanthology.org/2020.emnlp-main.367/",
    doi = "10.18653/v1/2020.emnlp-main.367",
    pages = "4536--4546"
}

@inproceedings{liang-etal-2023-xlm,
    title = "{XLM}-{V}: Overcoming the Vocabulary Bottleneck in Multilingual Masked Language Models",
    author = "Liang, Davis  and
      Gonen, Hila  and
      Mao, Yuning  and
      Hou, Rui  and
      Goyal, Naman  and
      Ghazvininejad, Marjan  and
      Zettlemoyer, Luke  and
      Khabsa, Madian",
    editor = "Bouamor, Houda  and
      Pino, Juan  and
      Bali, Kalika",
    booktitle = "Proceedings of the 2023 Conference on Empirical Methods in Natural Language Processing",
    month = dec,
    year = "2023",
    address = "Singapore",
    publisher = "Association for Computational Linguistics",
    url = "https://aclanthology.org/2023.emnlp-main.813/",
    doi = "10.18653/v1/2023.emnlp-main.813",
    pages = "13142--13152"
}

@inproceedings{wmt24-4african,
    title = "Correcting {FLORES} Evaluation Dataset for Four {A}frican Languages",
    author = "Abdulmumin, Idris  and
      Mkhwanazi, Sthembiso  and
      Mbooi, Mahlatse  and
      Muhammad, Shamsuddeen Hassan  and
      Ahmad, Ibrahim Said  and
      Putini, Neo  and
      Mathebula, Miehleketo  and
      Shingange, Matimba  and
      Gwadabe, Tajuddeen  and
      Marivate, Vukosi",
    editor = "Haddow, Barry  and
      Kocmi, Tom  and
      Koehn, Philipp  and
      Monz, Christof",
    booktitle = "Proceedings of the Ninth Conference on Machine Translation",
    month = nov,
    year = "2024",
    address = "Miami, Florida, USA",
    publisher = "Association for Computational Linguistics",
    url = "https://aclanthology.org/2024.wmt-1.44/",
    doi = "10.18653/v1/2024.wmt-1.44",
    pages = "570--578"
}

@book{Jolliffe2002,
  author    = {I. T. Jolliffe},
  title     = {Principal Component Analysis},
  edition   = {2nd},
  series    = {Springer Series in Statistics},
  year      = {2002},
  publisher = {Springer New York, NY},
  address   = {New York},
  doi       = {10.1007/b98835},
  isbn      = {978-0-387-95442-4}
}

@inproceedings{koto-etal-2024-zero,
    title = "Zero-shot Sentiment Analysis in Low-Resource Languages Using a Multilingual Sentiment Lexicon",
    author = "Koto, Fajri  and
      Beck, Tilman  and
      Talat, Zeerak  and
      Gurevych, Iryna  and
      Baldwin, Timothy",
    editor = "Graham, Yvette  and
      Purver, Matthew",
    booktitle = "Proceedings of the 18th Conference of the European Chapter of the Association for Computational Linguistics (Volume 1: Long Papers)",
    month = mar,
    year = "2024",
    address = "St. Julian{'}s, Malta",
    publisher = "Association for Computational Linguistics",
    url = "https://aclanthology.org/2024.eacl-long.18/",
    doi = "10.18653/v1/2024.eacl-long.18",
    pages = "298--320"
}

@inproceedings{wang-etal-2022-expanding,
    title = "Expanding Pretrained Models to Thousands More Languages via Lexicon-based Adaptation",
    author = "Wang, Xinyi  and
      Ruder, Sebastian  and
      Neubig, Graham",
    editor = "Muresan, Smaranda  and
      Nakov, Preslav  and
      Villavicencio, Aline",
    booktitle = "Proceedings of the 60th Annual Meeting of the Association for Computational Linguistics (Volume 1: Long Papers)",
    month = may,
    year = "2022",
    address = "Dublin, Ireland",
    publisher = "Association for Computational Linguistics",
    url = "https://aclanthology.org/2022.acl-long.61/",
    doi = "10.18653/v1/2022.acl-long.61",
    pages = "863--877"
}

@inproceedings{adebara-etal-2022-afrolid,
    title = "{A}fro{LID}: A Neural Language Identification Tool for {A}frican Languages",
    author = "Adebara, Ife  and
      Elmadany, AbdelRahim  and
      Abdul-Mageed, Muhammad  and
      Inciarte, Alcides",
    booktitle = "Proceedings of the 2022 Conference on Empirical Methods in Natural Language Processing",
    month = dec,
    year = "2022",
    address = "Abu Dhabi, United Arab Emirates",
    publisher = "Association for Computational Linguistics",
    url = "https://aclanthology.org/2022.emnlp-main.128/",
    doi = "10.18653/v1/2022.emnlp-main.128",
    pages = "1958--1981"
}

@inproceedings{muller-etal-2021-unseen,
    title = "When Being Unseen from m{BERT} is just the Beginning: Handling New Languages With Multilingual Language Models",
    author = "Muller, Benjamin  and
      Anastasopoulos, Antonios  and
      Sagot, Beno{\^i}t  and
      Seddah, Djam{\'e}",
    editor = "Toutanova, Kristina  and
      Rumshisky, Anna  and
      Zettlemoyer, Luke  and
      Hakkani-Tur, Dilek  and
      Beltagy, Iz  and
      Bethard, Steven  and
      Cotterell, Ryan  and
      Chakraborty, Tanmoy  and
      Zhou, Yichao",
    booktitle = "Proceedings of the 2021 Conference of the North American Chapter of the Association for Computational Linguistics: Human Language Technologies",
    month = jun,
    year = "2021",
    address = "Online",
    publisher = "Association for Computational Linguistics",
    url = "https://aclanthology.org/2021.naacl-main.38/",
    doi = "10.18653/v1/2021.naacl-main.38",
    pages = "448--462"
}

@inproceedings{ahia-etal-2023-languages,
    title = "Do All Languages Cost the Same? Tokenization in the Era of Commercial Language Models",
    author = "Ahia, Orevaoghene  and
      Kumar, Sachin  and
      Gonen, Hila  and
      Kasai, Jungo  and
      Mortensen, David  and
      Smith, Noah  and
      Tsvetkov, Yulia",
    editor = "Bouamor, Houda  and
      Pino, Juan  and
      Bali, Kalika",
    booktitle = "Proceedings of the 2023 Conference on Empirical Methods in Natural Language Processing",
    month = dec,
    year = "2023",
    address = "Singapore",
    publisher = "Association for Computational Linguistics",
    url = "https://aclanthology.org/2023.emnlp-main.614/",
    doi = "10.18653/v1/2023.emnlp-main.614",
    pages = "9904--9923"
}

\newpage
\appendix

\section{Language Resource Availability Details}
\label{sec:language_resource_availability}

Table~\ref{tab:language_resource_availability} presents the availability of language resources under study in the Wikipedia dumps. English is highlighted in lime, as it serves both as the reference for comparison to other lower-resource languages and the pivot language when constructing the parallel tokenizers, due to its large corpus size and richer vocabulary coverage.

\begin{table}[ht]
\centering
\footnotesize
\begin{tabular}{l|c|c|l|l}
\hline
\textbf{Lang.} & \textbf{iso} & \textbf{code} & \textbf{Script} & \textbf{Resource} \\
\hline
\cellcolor{lime!30}English & \cellcolor{lime!30}eng & \cellcolor{lime!30}[EN] & \cellcolor{lime!30}Latin & \cellcolor{lime!30}10.7GB \\
\hline
Acehnese & ace & [AC] & Latin & 2.52MB \\
Amharic & amh & [AM] & Amharic & 18MB \\
Balinese & ban & [BA] & Latin & 14.24MB \\
Hausa & hau & [HA] & Latin & 108MB \\
Igbo & ibo & [IG] & Latin & 104MB \\
Javanese & jav & [JV] & Latin & 50.7MB \\
Kinyarwanda & kin & [RW] & Latin & 12.25MB \\
Minangkabau & min & [MI] & Latin & 85.4MB \\
Oromo & orm & [OR] & Latin & 3.71MB \\
Sundanese & sun & [SU] & Latin & 34.1MB \\
Swahili & swa & [SW] & Latin & 54.4MB \\
Tigrinya & tir & [TI] & Amharic & 1.1MB \\
Twi & twi & [TW] & Latin & 8.6MB \\
\hline
\end{tabular}
\caption{Language resource availability. The `iso' column lists language codes following the ISO 639-3 convention, while the `code' column specifies the language identifiers used to determine the language embeddings in the model's input representation.}
\label{tab:language_resource_availability}
\end{table}

\section{Benchmark Language Coverage}
\label{sec:benchmark_language_coverage}

Table~\ref{tab:benchmark_language_coverage} reports the language coverage of each benchmark included in our study. For brevity, we use the abbreviations NusaX, Afri, Ethio, and BRIGHT to refer to the NusaX-senti, AfriHate, EthioEmo, and BRIGHTER datasets, respectively. The numbers next to the checkmarks indicate the number of instances used in the benchmarking process.

\begin{table}[htbp]
\centering
\small
\tabcolsep=0.2cm
\begin{tabular}{l|c|c|c|c}
\hline
\textbf{lang.} & \textbf{NusaX} & \textbf{Afri} & \textbf{Ethio} & \textbf{BRIGHT} \\
\hline
ace & \checkmark (1K) & & & \\ \hline
amh & & \checkmark (4.96K) & \checkmark (5.92K) & \\ \hline
ban & \checkmark (1K) & & & \\ \hline
hau & & \checkmark (6.64K) & & \checkmark (5.02K)\\ \hline
ibo & & \checkmark (5K) & & \checkmark (6.73K)\\ \hline
jav & \checkmark (1K) & & & \\ \hline
min & \checkmark (1K) & & & \\ \hline
kin & & \checkmark (4.72K) & & \checkmark (5.73K)\\ \hline
orm & & \checkmark (5.03K) & \checkmark (5.74K) & \\ \hline
sun & \checkmark (1K) & & & \checkmark (3.17K)\\ \hline
swa & & \checkmark (21.1K) & & \checkmark (7.72K) \\ \hline
tir & & \checkmark (5.07K) & \checkmark (6.14K) & \\ \hline
twi & & \checkmark (3.9K) & &\\
\hline
\end{tabular}
\caption{Benchmark language coverage.}
\label{tab:benchmark_language_coverage}
\end{table}

\section{Benchmarking Result Details}
\label{sec:apdx_main_result_detail}

Tables~\ref{tab:results_nusaxsenti_scratch}, \ref{tab:results_afrihate_scratch}, \ref{tab:results_ethioemo_scratch}, and \ref{tab:results_brighter_scratch} present the benchmarking results for NusaX-senti, AfriHate, EthioEmo, and BRIGHTER, respectively. Results are reported for each language in the benchmark under training data settings of 100\%, 50\%, 10\%, and 1\%. To ensure stability and reliability, each experiment was run three times with different random seeds, and the reported results include the corresponding standard deviations. Based on the average results, our method yields better results than the baseline, highlighting the performance of our method in cross-lingual settings.

\begin{table*}[tbp]
\centering
\small
\begin{tabular}{c|c|c|c|c|c|c|c}
\hline
\#data & Tokenizer & ace & ban & jav & min & sun & \textbf{avg} \\
\hline
\multirow{2}{*}{100\%} 
& Single-13L      &  \textbf{76.61 \textsubscript{($\pm$1.31)}} &  \textbf{73.23 \textsubscript{($\pm$1.00)}} & 77.43 \textsubscript{($\pm$1.45)} & 76.48 \textsubscript{($\pm$1.37)} & 76.68 \textsubscript{($\pm$1.19)} & 76.09 \textsubscript{($\pm$1.27)} \\
& Parallel-13L (ours)  & 74.22 \textsubscript{($\pm$1.91)} & 72.62 \textsubscript{($\pm$1.22)} &  \textbf{78.26 \textsubscript{($\pm$0.50)}} &  \textbf{78.43 \textsubscript{($\pm$0.42)}} &  \textbf{77.26 \textsubscript{($\pm$1.22)}} &  \textbf{76.16 \textsubscript{($\pm$1.05)}} \\
\hline
\multirow{2}{*}{50\%} 
& Single-13L      &  \textbf{73.44 \textsubscript{($\pm$1.41)}} & 69.68 \textsubscript{($\pm$1.76)} & 71.86 \textsubscript{($\pm$2.03)} & 73.14 \textsubscript{($\pm$1.01)} & 74.22 \textsubscript{($\pm$0.61)} & 72.47 \textsubscript{($\pm$1.36)} \\
& Parallel-13L (ours)   & 71.53 \textsubscript{($\pm$0.52)} &  \textbf{72.83 \textsubscript{($\pm$2.03)}} &  \textbf{74.34 \textsubscript{($\pm$0.86)}} &  \textbf{74.38 \textsubscript{($\pm$2.15)}} &  \textbf{74.52 \textsubscript{($\pm$0.31)}} &  \textbf{73.52 \textsubscript{($\pm$1.17)}} \\
\hline
\multirow{2}{*}{10\%} 
& Single-13L      & 63.63 \textsubscript{($\pm$2.35)} & 63.02 \textsubscript{($\pm$1.92)} &  \textbf{66.59 \textsubscript{($\pm$1.86)}} & 65.06 \textsubscript{($\pm$1.60)} & 65.51 \textsubscript{($\pm$1.46)} & 64.76 \textsubscript{($\pm$1.84)} \\
& Parallel-13L (ours)   &  \textbf{65.38 \textsubscript{($\pm$0.47)}} &  \textbf{65.07 \textsubscript{($\pm$1.38)}} & 65.65 \textsubscript{($\pm$5.20)} &  \textbf{68.62 \textsubscript{($\pm$1.19)}} &  \textbf{66.07 \textsubscript{($\pm$1.77)}} &  \textbf{66.16 \textsubscript{($\pm$2.00)}} \\
\hline
\multirow{2}{*}{1\%} 
& Single-13L      &  \textbf{32.98 \textsubscript{($\pm$3.42)}} & 30.46 \textsubscript{($\pm$5.47)} & 28.12 \textsubscript{($\pm$4.77)} & 34.90 \textsubscript{($\pm$2.78)} & 31.26 \textsubscript{($\pm$2.93)} & 31.54 \textsubscript{($\pm$3.87)} \\
& Parallel-13L (ours)   & 24.78 \textsubscript{($\pm$6.28)} &  \textbf{33.14 \textsubscript{($\pm$2.55)}} &  \textbf{39.18 \textsubscript{($\pm$6.41)}} &  \textbf{37.75 \textsubscript{($\pm$2.91)}} &  \textbf{34.29 \textsubscript{($\pm$1.34)}} &  \textbf{33.83 \textsubscript{($\pm$3.90)}} \\
\hline
\end{tabular}
\caption{NusaX-senti benchmarking results. `ace', `ban', `jav', `min', and `sun' denote Acehnese, Balinese, Javanese, Minangkabau, and Sundanese, respectively.}
\label{tab:results_nusaxsenti_scratch}
\end{table*}

\begin{table*}[tbp]
\centering
\tiny
\tabcolsep=0.13cm
\begin{tabular}{c|c|c|c|c|c|c|c|c|c|c}
\hline
\#data & Tokenizer & amh & hau & ibo & kin & orm & swa & tir & twi & \textbf{avg} \\
\hline
\multirow{2}{*}{100\%} 
& Single-13L      & 59.72 \textsubscript{($\pm$1.80)} & 69.51 \textsubscript{($\pm$1.94)} & 85.36 \textsubscript{($\pm$0.74)} & 72.96 \textsubscript{($\pm$1.54)} &  \textbf{62.21 \textsubscript{($\pm$0.73)}} &  \textbf{90.29 \textsubscript{($\pm$0.39)}} & 60.91 \textsubscript{($\pm$1.09)} &  \textbf{55.91 \textsubscript{($\pm$1.27)}} & 69.61 \textsubscript{($\pm$1.19)} \\
& Parallel-13L (ours)   &  \textbf{59.91 \textsubscript{($\pm$0.35)}} &  \textbf{70.93 \textsubscript{($\pm$2.19)}} &  \textbf{87.62 \textsubscript{($\pm$0.85)}} &  \textbf{73.44 \textsubscript{($\pm$1.56)}} & 61.45 \textsubscript{($\pm$0.59)} & 89.94 \textsubscript{($\pm$0.22)} &  \textbf{61.61 \textsubscript{($\pm$1.07)}} & 53.47 \textsubscript{($\pm$2.77)} &  \textbf{69.80 \textsubscript{($\pm$1.20)}} \\
\hline
\multirow{2}{*}{50\%} 
& Single-13L      & 55.15 \textsubscript{($\pm$1.98)} & 66.23 \textsubscript{($\pm$1.76)} & 84.30 \textsubscript{($\pm$0.20)} &  \textbf{71.21 \textsubscript{($\pm$1.57)}} & 59.44 \textsubscript{($\pm$0.77)} & 87.41 \textsubscript{($\pm$0.46)} &  \textbf{61.33 \textsubscript{($\pm$1.21)}} &  \textbf{53.01 \textsubscript{($\pm$2.95)}} & 67.26 \textsubscript{($\pm$1.36)} \\
& Parallel-13L (ours)   &  \textbf{57.23 \textsubscript{($\pm$0.65)}} &  \textbf{67.89 \textsubscript{($\pm$2.15)}} &  \textbf{85.47 \textsubscript{($\pm$0.67)}} & 69.39 \textsubscript{($\pm$1.45)} &  \textbf{60.16 \textsubscript{($\pm$0.86)}} &  \textbf{88.60 \textsubscript{($\pm$0.08)}} & 59.50 \textsubscript{($\pm$0.76)} & 50.96 \textsubscript{($\pm$0.92)} &  \textbf{67.40 \textsubscript{($\pm$0.94)}} \\
\hline
\multirow{2}{*}{10\%} 
& Single-13L      & 50.83 \textsubscript{($\pm$0.29)} & 61.34 \textsubscript{($\pm$0.87)} & 77.39 \textsubscript{($\pm$1.80)} & 56.89 \textsubscript{($\pm$1.52)} & 50.49 \textsubscript{($\pm$6.57)} & 82.78 \textsubscript{($\pm$1.13)} &  \textbf{51.52 \textsubscript{($\pm$2.53)}} &  \textbf{47.13 \textsubscript{($\pm$1.98)}} & 59.80 \textsubscript{($\pm$2.09)} \\
& Parallel-13L (ours)   &  \textbf{51.27 \textsubscript{($\pm$2.07)}} &  \textbf{62.50 \textsubscript{($\pm$1.78)}} &  \textbf{79.93 \textsubscript{($\pm$1.09)}} &  \textbf{63.93 \textsubscript{($\pm$0.15)}} &  \textbf{51.08 \textsubscript{($\pm$0.38)}} & 84.24 \textsubscript{($\pm$1.19)} & 49.10 \textsubscript{($\pm$1.47)} & 39.29 \textsubscript{($\pm$3.87)} &  \textbf{60.17 \textsubscript{($\pm$1.50)}} \\
\hline
\multirow{2}{*}{1\%} 
& Single-13L      & 41.41 \textsubscript{($\pm$2.79)} &  \textbf{43.31 \textsubscript{($\pm$7.82)}} & 47.25 \textsubscript{($\pm$6.60)} & 42.37 \textsubscript{($\pm$3.92)} & 41.98 \textsubscript{($\pm$4.11)} & 69.11 \textsubscript{($\pm$2.53)} &  \textbf{41.49 \textsubscript{($\pm$4.06)}} & 30.70 \textsubscript{($\pm$3.79)} & 44.70 \textsubscript{($\pm$4.45)} \\
& Parallel-13L (ours)   &  \textbf{42.70 \textsubscript{($\pm$1.03)}} & 43.09 \textsubscript{($\pm$3.84)} &  \textbf{55.44 \textsubscript{($\pm$3.25)}} &  \textbf{48.10 \textsubscript{($\pm$1.81)}} & 42.35 \textsubscript{($\pm$3.00)} &  \textbf{71.33 \textsubscript{($\pm$2.22)}} & 39.56 \textsubscript{($\pm$2.28)} &  \textbf{31.26 \textsubscript{($\pm$2.74)}} &  \textbf{46.73 \textsubscript{($\pm$2.52)}} \\
\hline
\end{tabular}
\caption{AfriHate benchmarking results. `amh', `hau', `ibo', `kin', `orm', `swh', `tir', and `twi' denote Amharic, Hausa, Igbo, Kinyarwanda, Oromo, Swahili, Tigrinya, and Twi, respectively.}
\label{tab:results_afrihate_scratch}
\end{table*}

\begin{table*}[tbp]
\centering
\small
\begin{tabular}{c|c|c|c|c|c}
\hline
\#data & Tokenizer & amh & orm & tir & \textbf{avg} \\
\hline
\multirow{2}{*}{100\%} 
& Single-13L      & 55.07 \textsubscript{($\pm$1.01)} & 58.96 \textsubscript{($\pm$1.18)} & 50.92 \textsubscript{($\pm$0.75)} & 54.98 \textsubscript{($\pm$0.98)} \\
& Parallel-13L (ours)   &  \textbf{60.22 \textsubscript{($\pm$0.32)}} & \textbf{59.59 \textsubscript{($\pm$1.18)}} &  \textbf{51.21 \textsubscript{($\pm$1.03)}} &  \textbf{57.01 \textsubscript{($\pm$0.84)}} \\
\hline
\multirow{2}{*}{50\%} 
& Single-13L      & 52.98 \textsubscript{($\pm$2.09)} & 55.53 \textsubscript{($\pm$1.06)} & 45.48 \textsubscript{($\pm$1.31)} & 51.33 \textsubscript{($\pm$1.48)} \\
& Parallel-13L (ours)   &  \textbf{58.33 \textsubscript{($\pm$1.12)}} & \textbf{55.64 \textsubscript{($\pm$1.78)}} &  \textbf{48.86 \textsubscript{($\pm$2.44)}} &  \textbf{54.27 \textsubscript{($\pm$1.78)}} \\
\hline
\multirow{2}{*}{10\%} 
& Single-13L      &  \textbf{44.32 \textsubscript{($\pm$2.86)}} & 43.18 \textsubscript{($\pm$1.03)} &  \textbf{38.79 \textsubscript{($\pm$2.66)}} &  \textbf{42.10 \textsubscript{($\pm$2.19)}} \\
& Parallel-13L (ours)   & 44.00 \textsubscript{($\pm$2.86)} &  \textbf{43.36 \textsubscript{($\pm$1.73)}} & 36.68 \textsubscript{($\pm$2.04)} & 41.35 \textsubscript{($\pm$2.21)} \\
\hline
\multirow{2}{*}{1\%} 
& Single-13L      &  \textbf{34.85 \textsubscript{($\pm$0.34)}} & 24.37 \textsubscript{($\pm$3.50)} &  \textbf{22.91 \textsubscript{($\pm$1.81)}} &  \textbf{27.37 \textsubscript{($\pm$1.88)}} \\
& Parallel-13L (ours)   & 33.77 \textsubscript{($\pm$1.83)} &  \textbf{24.42 \textsubscript{($\pm$1.67)}} & 21.55 \textsubscript{($\pm$0.62)} & 26.58 \textsubscript{($\pm$1.37)} \\
\hline
\end{tabular}
\caption{EthioEmo benchmarking results. `amh', `orm', and `tir' denote Amharic, Oromo, and Tigrinya, respectively.}
\label{tab:results_ethioemo_scratch}
\end{table*}

\begin{table*}[tbp]
\centering
\small
\begin{tabular}{c|c|c|c|c|c|c|c}
\hline
\#data & Tokenizer & hau & ibo & kin & sun & swa & avg \\
\hline
\multirow{2}{*}{100\%} 
& Single-13L      & 58.66 \textsubscript{($\pm$0.16)} &  \textbf{58.72 \textsubscript{($\pm$0.63)}} & 41.65 \textsubscript{($\pm$0.58)} & 60.11 \textsubscript{($\pm$0.61)} & 22.16 \textsubscript{($\pm$2.71)} & 48.26 \textsubscript{($\pm$0.94)} \\
& Parallel-13L (ours)   &  \textbf{60.20 \textsubscript{($\pm$1.96)}} & 58.37 \textsubscript{($\pm$1.57)} &  \textbf{45.41 \textsubscript{($\pm$2.10)}} & \textbf{60.49 \textsubscript{($\pm$1.05)}} & \textbf{23.95 \textsubscript{($\pm$2.76)}} &  \textbf{49.68 \textsubscript{($\pm$1.89)}} \\
\hline
\multirow{2}{*}{50\%} 
& Single-13L      & 53.89 \textsubscript{($\pm$1.29)} &  \textbf{55.08 \textsubscript{($\pm$0.58)}} & 39.36 \textsubscript{($\pm$1.83)} &  \textbf{59.77 \textsubscript{($\pm$1.05)}} & 20.76 \textsubscript{($\pm$2.52)} & 45.77 \textsubscript{($\pm$1.46)} \\
& Parallel-13L (ours)   &  \textbf{56.65 \textsubscript{($\pm$0.81)}} & 55.05 \textsubscript{($\pm$1.65)} &  \textbf{43.42 \textsubscript{($\pm$0.81)}} & 57.56 \textsubscript{($\pm$2.02)} & \textbf{21.13 \textsubscript{($\pm$2.26)}} &  \textbf{46.76 \textsubscript{($\pm$1.51)}} \\
\hline
\multirow{2}{*}{10\%} 
& Single-13L      & 43.03 \textsubscript{($\pm$1.65)} &  \textbf{45.12 \textsubscript{($\pm$2.80)}} & 27.62 \textsubscript{($\pm$0.60)} & 50.63 \textsubscript{($\pm$4.83)} & \textbf{16.78 \textsubscript{($\pm$1.25)}} & 36.64 \textsubscript{($\pm$2.23)} \\
& Parallel-13L (ours)   &  \textbf{45.73 \textsubscript{($\pm$0.70)}} & 45.00 \textsubscript{($\pm$2.27)} &  \textbf{33.12 \textsubscript{($\pm$0.68)}} & \textbf{51.95 \textsubscript{($\pm$2.11)}} & 16.74 \textsubscript{($\pm$0.58)} &  \textbf{38.51 \textsubscript{($\pm$1.27)}} \\
\hline
\multirow{2}{*}{1\%} 
& Single-13L      & 9.02 \textsubscript{($\pm$8.00)} &  \textbf{21.28 \textsubscript{($\pm$3.23)}} & 14.89 \textsubscript{($\pm$0.76)} & 42.50 \textsubscript{($\pm$0.84)} & 0.00 \textsubscript{($\pm$0.00)} & 17.54 \textsubscript{($\pm$2.57)} \\
& Parallel-13L (ours)   &  \textbf{14.91 \textsubscript{($\pm$3.24)}} & 15.97 \textsubscript{($\pm$13.91)} &  \textbf{17.20 \textsubscript{($\pm$1.93)}} &  \textbf{46.30 \textsubscript{($\pm$0.72)}} & 0.00 \textsubscript{($\pm$0.00)} &  \textbf{18.88 \textsubscript{($\pm$3.96)}} \\
\hline
\end{tabular}
\caption{BRIGHTER benchmarking results. `hau', `ibo', `kin', `sun', and `swa' denote Hausa, Igbo, Kinyarwanda, Sundanese, and Swahili, respectively.}
\label{tab:results_brighter_scratch}
\end{table*}

\section{Bitext Mining Details}
\label{sec:bitext_mining_details}

Tables~\ref{tab:bitext_mining_3} and \ref{tab:bitext_mining_5} present the bitext mining performance of each model using their corresponding tokenizers: \textit{Single-13L} and \textit{Parallel-13L}. Performance is evaluated using the \texttt{xsim} error rate, which measures how accurately a model selects the equivalent sentence across languages based on the FLORES+ parallel data. A lower error rate indicates better cross-lingual alignment, making this metric an effective indicator of a model's ability to capture cross-lingual representations.

\begin{table*}[tbp]
\centering
\small
\begin{tabular}{|l|c|c|c|c|c|c|c|c|c|c|c|c|c|}
\hline
 & ace & amh & ban & hau & ibo & jav & min & kin & orm & sun & swa & tir & twi \\
\hline
ace & \cellcolor{black!50} & \textbf{97.83} & 82.21 & 85.67 & 88.34 & 73.12 & \textbf{72.83} & \textbf{91.01} & 97.04 & \textbf{77.57} & 86.26 & \textbf{98.52} & \textbf{89.62} \\
\hline
amh & \cellcolor{black!50} & \cellcolor{black!50} & 96.74 & 96.34 & 96.94 & 96.44 & 96.74 & 97.53 & 97.73 & 96.94 & 96.54 & 90.51 & 98.02 \\
\hline
ban & \cellcolor{black!50} & \cellcolor{black!50} & \cellcolor{black!50} & 72.83 & 77.87 & \textbf{35.97} & 48.02 & 84.19 & 96.34 & 47.43 & 71.54 & 98.22 & 85.28 \\
\hline
hau & \cellcolor{black!50} & \cellcolor{black!50} & \cellcolor{black!50} & \cellcolor{black!50} & 64.43 & 62.15 & 72.13 & 82.91 & 95.26 & 68.38 & 63.54 & 97.63 & 81.72 \\
\hline
ibo & \cellcolor{black!50} & \cellcolor{black!50} & \cellcolor{black!50} & \cellcolor{black!50} & \cellcolor{black!50} & 70.95 & 78.46 & 87.15 & 96.84 & 75.69 & 71.05 & 98.02 & 85.97 \\
\hline
jav & \cellcolor{black!50} & \cellcolor{black!50} & \cellcolor{black!50} & \cellcolor{black!50} & \cellcolor{black!50} & \cellcolor{black!50} & 43.58 & 85.38 & 96.44 & \textbf{34.88} & 63.64 & 98.42 & 83.70 \\
\hline
min & \cellcolor{black!50} & \cellcolor{black!50} & \cellcolor{black!50} & \cellcolor{black!50} & \cellcolor{black!50} & \cellcolor{black!50} & \cellcolor{black!50} & 84.19 & 96.34 & \textbf{42.89} & 69.76 & 98.32 & 84.09 \\
\hline
kin & \cellcolor{black!50} & \cellcolor{black!50} & \cellcolor{black!50} & \cellcolor{black!50} & \cellcolor{black!50} & \cellcolor{black!50} & \cellcolor{black!50} & \cellcolor{black!50} & 97.04 & \textbf{84.68} & 78.36 & 98.32 & 88.54 \\
\hline
orm & \cellcolor{black!50} & \cellcolor{black!50} & \cellcolor{black!50} & \cellcolor{black!50} & \cellcolor{black!50} & \cellcolor{black!50} & \cellcolor{black!50} & \cellcolor{black!50} & \cellcolor{black!50} & 97.23 & 96.64 & \textbf{98.12} & \textbf{96.84} \\
\hline
sun & \cellcolor{black!50} & \cellcolor{black!50} & \cellcolor{black!50} & \cellcolor{black!50} & \cellcolor{black!50} & \cellcolor{black!50} & \cellcolor{black!50} & \cellcolor{black!50} & \cellcolor{black!50} & \cellcolor{black!50} & 67.29 & 98.62 & 84.58 \\
\hline
swa & \cellcolor{black!50} & \cellcolor{black!50} & \cellcolor{black!50} & \cellcolor{black!50} & \cellcolor{black!50} & \cellcolor{black!50} & \cellcolor{black!50} & \cellcolor{black!50} & \cellcolor{black!50} & \cellcolor{black!50} & \cellcolor{black!50} & 98.52 & 82.11 \\
\hline
tir & \cellcolor{black!50} & \cellcolor{black!50} & \cellcolor{black!50} & \cellcolor{black!50} & \cellcolor{black!50} & \cellcolor{black!50} & \cellcolor{black!50} & \cellcolor{black!50} & \cellcolor{black!50} & \cellcolor{black!50} & \cellcolor{black!50} & \cellcolor{black!50} & 98.91 \\
\hline
twi & \cellcolor{black!50} & \cellcolor{black!50} & \cellcolor{black!50} & \cellcolor{black!50} & \cellcolor{black!50} & \cellcolor{black!50} & \cellcolor{black!50} & \cellcolor{black!50} & \cellcolor{black!50} & \cellcolor{black!50} & \cellcolor{black!50} & \cellcolor{black!50} & \cellcolor{black!50} \\
\hline
\end{tabular}
\caption{\textit{Single-13L} bitext mining scores, calculated using \texttt{xsim} error rate score.}
\label{tab:bitext_mining_3}
\end{table*}

\begin{table*}[tbp]
\centering
\small
\begin{tabular}{|l|c|c|c|c|c|c|c|c|c|c|c|c|c|}
\hline
 & ace & amh & ban & hau & ibo & jav & min & kin & orm & sun & swa & tir & twi \\
\hline
ace & \cellcolor{black!50} & 98.32 & 88.04 & \textbf{76.98} & \textbf{85.77} & \textbf{71.34} & 81.23 & 98.62 & 97.92 & 82.31 & \textbf{70.36} & 98.81 & 98.22 \\
\hline
amh & \cellcolor{black!50} & \cellcolor{black!50} & \textbf{77.87} & \textbf{73.62} & \textbf{85.08} & \textbf{69.96} & \textbf{72.43} & \textbf{83.30} & \textbf{95.45} & \textbf{81.92} & \textbf{65.61} & 97.33 & \textbf{86.96} \\
\hline
ban & \cellcolor{black!50} & \cellcolor{black!50} & \cellcolor{black!50} & \textbf{57.11} & \textbf{68.18} & 50.79 & \textbf{46.25} & \textbf{70.95} & \textbf{85.87} & \textbf{46.25} & \textbf{47.73} & \textbf{95.45} & \textbf{76.48} \\
\hline
hau & \cellcolor{black!50} & \cellcolor{black!50} & \cellcolor{black!50} & \cellcolor{black!50} & \textbf{52.87} & \textbf{35.57} & \textbf{41.40} & \textbf{60.18} & \textbf{82.91} & \textbf{57.11} & \textbf{30.93} & \textbf{89.43} & \textbf{70.55} \\
\hline
ibo & \cellcolor{black!50} & \cellcolor{black!50} & \cellcolor{black!50} & \cellcolor{black!50} & \cellcolor{black!50} & \textbf{58.10} & \textbf{68.28} & \textbf{78.36} & \textbf{87.35} & \textbf{72.53} & \textbf{54.25} & \textbf{97.23} & \textbf{77.27} \\
\hline
jav & \cellcolor{black!50} & \cellcolor{black!50} & \cellcolor{black!50} & \cellcolor{black!50} & \cellcolor{black!50} & \cellcolor{black!50} & \textbf{22.92} & \textbf{56.92} & \textbf{79.84} & 44.27 & \textbf{24.51} & \textbf{92.98} & \textbf{75.30} \\
\hline
min & \cellcolor{black!50} & \cellcolor{black!50} & \cellcolor{black!50} & \cellcolor{black!50} & \cellcolor{black!50} & \cellcolor{black!50} & \cellcolor{black!50} & \textbf{60.77} & \textbf{79.45} & 87.15 & \textbf{29.74} & \textbf{95.16} & \textbf{71.44} \\
\hline
kin & \cellcolor{black!50} & \cellcolor{black!50} & \cellcolor{black!50} & \cellcolor{black!50} & \cellcolor{black!50} & \cellcolor{black!50} & \cellcolor{black!50} & \cellcolor{black!50} & \textbf{92.29} & 89.03 & \textbf{59.19} & \textbf{96.15} & \textbf{81.92} \\
\hline
orm & \cellcolor{black!50} & \cellcolor{black!50} & \cellcolor{black!50} & \cellcolor{black!50} & \cellcolor{black!50} & \cellcolor{black!50} & \cellcolor{black!50} & \cellcolor{black!50} & \cellcolor{black!50} & \textbf{93.08} & \textbf{83.70} & 98.62 & 97.23 \\
\hline
sun & \cellcolor{black!50} & \cellcolor{black!50} & \cellcolor{black!50} & \cellcolor{black!50} & \cellcolor{black!50} & \cellcolor{black!50} & \cellcolor{black!50} & \cellcolor{black!50} & \cellcolor{black!50} & \cellcolor{black!50} & \textbf{47.33} & \textbf{92.89} & \textbf{72.92} \\
\hline
swa & \cellcolor{black!50} & \cellcolor{black!50} & \cellcolor{black!50} & \cellcolor{black!50} & \cellcolor{black!50} & \cellcolor{black!50} & \cellcolor{black!50} & \cellcolor{black!50} & \cellcolor{black!50} & \cellcolor{black!50} & \cellcolor{black!50} & \textbf{87.65} & \textbf{70.36} \\
\hline
tir & \cellcolor{black!50} & \cellcolor{black!50} & \cellcolor{black!50} & \cellcolor{black!50} & \cellcolor{black!50} & \cellcolor{black!50} & \cellcolor{black!50} & \cellcolor{black!50} & \cellcolor{black!50} & \cellcolor{black!50} & \cellcolor{black!50} & \cellcolor{black!50} & \textbf{98.62} \\
\hline
twi & \cellcolor{black!50} & \cellcolor{black!50} & \cellcolor{black!50} & \cellcolor{black!50} & \cellcolor{black!50} & \cellcolor{black!50} & \cellcolor{black!50} & \cellcolor{black!50} & \cellcolor{black!50} & \cellcolor{black!50} & \cellcolor{black!50} & \cellcolor{black!50} & \cellcolor{black!50} \\
\hline
\end{tabular}
\caption{\textit{Parallel-13L} bitext mining scores, calculated using \texttt{xsim} error rate score.}
\label{tab:bitext_mining_5}
\end{table*}

\section{Cross-Lingual Transfer under Limited Target-Language Data Details}
\label{sec:cross_lingual_auxiliary}

Similar to Appendix~\ref{sec:apdx_main_result_detail}, we report benchmarking results under a limited target-language data setting, where the auxiliary languages are fully utilized (100\% of the data) during finetuning, while the target language is limited to 50\% or 0\% of its data. This setup highlights cross-lingual transfer from the auxiliary languages to the target language. Tables~\ref{tab:results_nusaxsenti_crosslingual_scratch}, \ref{tab:results_afrihate_crosslingual_scratch}, \ref{tab:results_ethioemo_crosslingual_scratch}, and \ref{tab:results_brigher_crosslingual_scratch} present the results for NusaX-senti, AfriHate, EthioEmo, and BRIGHTER, respectively. For each language, we repeat the experiments three times with different seeds to ensure robustness.

\begin{table*}[tbp]
\centering
\small
\begin{tabular}{c|c|c|c|c|c|c|c}
\hline
\#data & Tokenizer & ace & ban & jav & min & sun & avg \\
\hline
\multirow{2}{*}{50\%} 
& Single-13L  &  \textbf{75.99 \textsubscript{($\pm$0.88)}} & 70.69 \textsubscript{($\pm$2.97)} &  \textbf{78.00 \textsubscript{($\pm$0.82)}} & 74.92 \textsubscript{($\pm$2.56)} &  \textbf{76.50 \textsubscript{($\pm$0.72)}} & 75.22 \textsubscript{($\pm$1.59)} \\
& Parallel-13L (ours)  & 73.75 \textsubscript{($\pm$0.62)} &  \textbf{71.94 \textsubscript{($\pm$1.05)}} & 77.01 \textsubscript{($\pm$1.52)} &  \textbf{78.15 \textsubscript{($\pm$0.70)}} & 76.22 \textsubscript{($\pm$2.10)} &  \textbf{75.42 \textsubscript{($\pm$1.20)}} \\
\hline
\multirow{2}{*}{0\%} 
& Single-13L  & 66.77 \textsubscript{($\pm$2.59)} &  \textbf{69.76 \textsubscript{($\pm$2.20)}} & 73.93 \textsubscript{($\pm$2.37)} & 74.09 \textsubscript{($\pm$1.85)} &  \textbf{74.60 \textsubscript{($\pm$0.95)}} & 71.83 \textsubscript{($\pm$1.99)} \\
& Parallel-13L (ours)  &  \textbf{69.44 \textsubscript{($\pm$1.94)}} & 67.43 \textsubscript{($\pm$1.27)} &  \textbf{76.66 \textsubscript{($\pm$0.71)}} &  \textbf{75.87 \textsubscript{($\pm$2.05)}} & 74.12 \textsubscript{($\pm$1.50)} &  \textbf{72.70 \textsubscript{($\pm$1.49)}} \\
\hline
\end{tabular}
\caption{NusaX-senti benchmarking results under a limited target-language data setting. `ace', `ban', `jav', `min', and `sun' denote Acehnese, Balinese, Javanese, Minangkabau, and Sundanese, respectively.}
\label{tab:results_nusaxsenti_crosslingual_scratch}
\end{table*}

\begin{table*}[tbp]
\centering
\tiny
\tabcolsep=0.13cm
\begin{tabular}{c|c|c|c|c|c|c|c|c|c|c}
\hline
\#data & Tokenizer & amh & hau & ibo & kin & orm & swh & tir & twi & avg \\
\hline
\multirow{2}{*}{50\%} 
& Single-13L  &  \textbf{57.92 \textsubscript{($\pm$0.69)}} & 65.35 \textsubscript{($\pm$1.71)} & 84.20 \textsubscript{($\pm$0.43)} & 59.20 \textsubscript{($\pm$1.52)} & 69.23 \textsubscript{($\pm$0.80)} & 88.22 \textsubscript{($\pm$0.55)} & 57.75 \textsubscript{($\pm$3.59)} & 54.63 \textsubscript{($\pm$2.30)} &  \textbf{67.06 \textsubscript{($\pm$1.45)}} \\
& Parallel-13L (ours)  & 57.01 \textsubscript{($\pm$0.13)} & 66.18 \textsubscript{($\pm$1.73)} &  \textbf{85.27 \textsubscript{($\pm$1.52)}} &  \textbf{59.35 \textsubscript{($\pm$2.04)}} &  \textbf{70.77 \textsubscript{($\pm$0.62)}} &  \textbf{88.73 \textsubscript{($\pm$0.11)}} &  \textbf{58.57 \textsubscript{($\pm$1.41)}} & 50.13 \textsubscript{($\pm$3.01)} & 67.00 \textsubscript{($\pm$1.32)} \\
\hline
\multirow{2}{*}{0\%} 
& Single-13L  & 42.32 \textsubscript{($\pm$2.74)} & 39.32 \textsubscript{($\pm$3.11)} & 38.91 \textsubscript{($\pm$1.91)} & 36.86 \textsubscript{($\pm$0.78)} & 40.06 \textsubscript{($\pm$4.86)} & 42.06 \textsubscript{($\pm$0.67)} & 35.27 \textsubscript{($\pm$2.89)} & 23.37 \textsubscript{($\pm$4.55)} & 37.27 \textsubscript{($\pm$2.69)} \\
& Parallel-13L (ours)  &  \textbf{46.38 \textsubscript{($\pm$2.74)}} &  \textbf{45.13 \textsubscript{($\pm$3.04)}} &  \textbf{45.13 \textsubscript{($\pm$4.18)}} &  \textbf{41.32 \textsubscript{($\pm$1.07)}} &  \textbf{44.78 \textsubscript{($\pm$4.19)}} &  \textbf{44.11 \textsubscript{($\pm$5.10)}} & 32.45 \textsubscript{($\pm$2.41)} &  \textbf{26.08 \textsubscript{($\pm$1.68)}} &  \textbf{40.67 \textsubscript{($\pm$3.05)}} \\
\hline
\end{tabular}
\caption{AfriHate benchmarking results under a limited target-language data setting. `amh', `hau', `ibo', `kin', `orm', `swh', `tir', and `twi' denote Amharic, Hausa, Igbo, Kinyarwanda, Oromo, Swahili, Tigrinya, and Twi, respectively.}
\label{tab:results_afrihate_crosslingual_scratch}
\end{table*}

\begin{table*}[tbp]
\centering
\small
\begin{tabular}{c|c|c|c|c|c}
\hline
\#data & Tokenizer & amh & orm & tir & avg \\
\hline
\multirow{2}{*}{50\%} 
& Single-13L  & 53.96 \textsubscript{($\pm$1.26)} & 54.65 \textsubscript{($\pm$1.48)} & 46.60 \textsubscript{($\pm$0.91)} & 51.73 \textsubscript{($\pm$1.22)} \\
& Parallel-13L (ours)  &  \textbf{58.39 \textsubscript{($\pm$0.51)}} &  \textbf{55.77 \textsubscript{($\pm$0.82)}} &  \textbf{46.74 \textsubscript{($\pm$3.27)}} &  \textbf{53.63 \textsubscript{($\pm$1.53)}} \\
\hline
\multirow{2}{*}{0\%} 
& Single-13L  &  \textbf{32.58 \textsubscript{($\pm$4.21)}} & 12.10 \textsubscript{($\pm$4.66)} &  \textbf{26.55 \textsubscript{($\pm$0.65)}} &  \textbf{23.75 \textsubscript{($\pm$3.17)}} \\
& Parallel-13L (ours)  & 27.46 \textsubscript{($\pm$3.60)} &  \textbf{18.61 \textsubscript{($\pm$0.27)}} & 23.53 \textsubscript{($\pm$4.77)} & 23.20 \textsubscript{($\pm$2.88)} \\
\hline
\end{tabular}
\caption{EthioEmo benchmarking results under a limited target-language data setting. `amh', `orm', and `tir' denote Amharic, Oromo, and Tigrinya, respectively.}
\label{tab:results_ethioemo_crosslingual_scratch}
\end{table*}

\begin{table*}[tbp]
\centering
\small
\begin{tabular}{c|c|c|c|c|c|c|c}
\hline
\#data & Tokenizer & hau & ibo & kin & sun & swa & avg \\
\hline
\multirow{2}{*}{50\%} 
& Single-13L  & 52.92 \textsubscript{($\pm$1.09)} & 55.41 \textsubscript{($\pm$0.32)} & 39.57 \textsubscript{($\pm$2.08)} & 58.18 \textsubscript{($\pm$1.78)} & 18.95 \textsubscript{($\pm$1.33)} & 45.00 \textsubscript{($\pm$1.32)} \\
& Parallel-13L (ours)  &  \textbf{57.36 \textsubscript{($\pm$0.62)}} &  \textbf{56.64 \textsubscript{($\pm$0.95)}} &  \textbf{41.72 \textsubscript{($\pm$0.83)}} & 57.89 \textsubscript{($\pm$1.38)} &  \textbf{21.96 \textsubscript{($\pm$1.39)}} &  \textbf{47.11 \textsubscript{($\pm$1.04)}} \\
\hline
\multirow{2}{*}{0\%} 
& Single-3L  & 15.13 \textsubscript{($\pm$3.30)} &  \textbf{16.31 \textsubscript{($\pm$0.61)}} & 9.72 \textsubscript{($\pm$0.59)} &  \textbf{18.93 \textsubscript{($\pm$2.24)}} & 15.13 \textsubscript{($\pm$1.45)} & 15.04 \textsubscript{($\pm$1.64)} \\
& Parallel-13L (ours)  &  \textbf{26.92 \textsubscript{($\pm$2.65)}} & 16.08 \textsubscript{($\pm$2.15)} &  \textbf{23.37 \textsubscript{($\pm$0.32)}} & 15.90 \textsubscript{($\pm$0.96)} & 16.38 \textsubscript{($\pm$3.31)} &  \textbf{19.73 \textsubscript{($\pm$1.88)}} \\
\hline
\end{tabular}
\caption{BRIGHTER benchmarking results under a limited target-language data setting. `hau', `ibo', `kin', `sun', and `swa' denote Hausa, Igbo, Kinyarwanda, Sundanese, and Swahili, respectively.}
\label{tab:results_brigher_crosslingual_scratch}
\end{table*}


\begin{figure}[t]
\centering
  \includegraphics[width=0.8\columnwidth]{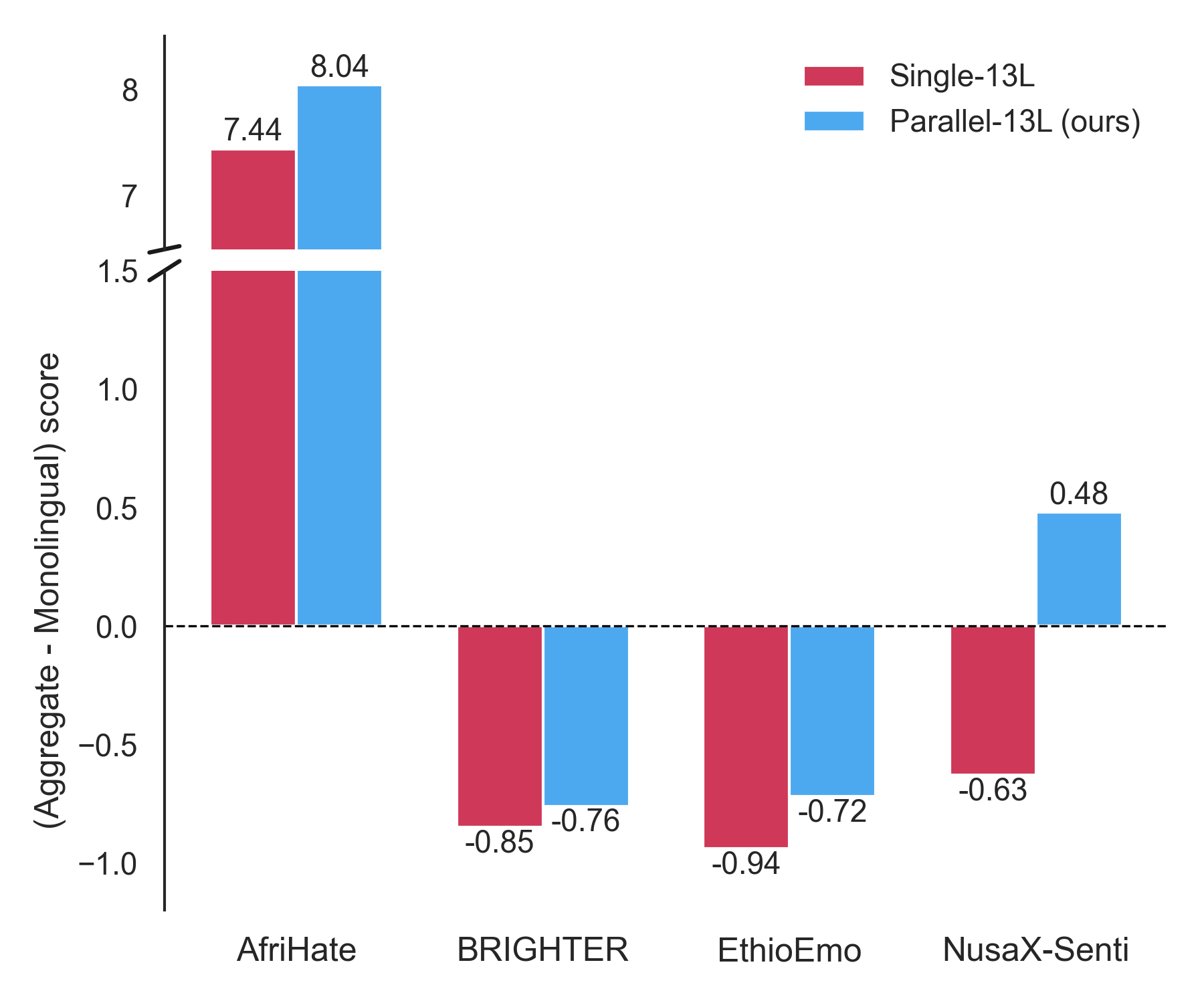}
  \caption{Performance differences across benchmarks on both cross-lingual and monolingual finetuning data for models pretrained with different tokenizer setups.}
  \label{fig:aggr_minus_mono}
\end{figure}

\subsection{How Much Does Multilingual Training Improve Over Single-Language Training?}

This analysis compares fine-tuning on concatenated multilingual datasets versus training solely on a target language's data. As shown in Figure~\ref{fig:aggr_minus_mono}, both baselines and our method benefit from multilingual training in the AfriHate benchmark, with average gains of 7-8\%. In contrast, improvements are smaller for BRIGHTER and EthioEmo, while in NusaX-senti, our method still achieves a modest 0.48\% increase. Between \textit{Single-13L} and \textit{Parallel-13L}, the latter benefits more consistently from multilingual concatenation, suggesting that it leverages shared cross-lingual information more effectively. Detailed results for monolingual fine-tuning are provided in Appendix~\ref{sec:details_monolingual}, showing that overall performance is comparable, though \textit{Parallel-13L} maintains slightly higher scores on average.

\section{Monolingual Benchmarking Details}
\label{sec:details_monolingual}

The benchmarking results using monolingual training data are shown in Tables~\ref{tab:nusaxsenti_mono_scratch}, \ref{tab:afrihate_mono_scratch}, \ref{tab:ethioemo_mono_scratch}, and \ref{tab:brighter_mono_scratch}. Each experiment was repeated three times with preselected seeds, and we report the averages along with the standard deviations for each language. Overall, the results are comparable: \textit{Single-13L} performs better on the NusaX-senti and AfriHate benchmarks, while \textit{Parallel-13L} performs better on emotion classification tasks such as EthioEmo and BRIGHTER. On average across all benchmarks, however, \textit{Parallel-13L} outperforms \textit{Single-13L} by 0.42\%.

\begin{table*}[tbp]
\centering
\small
\begin{tabular}{c|c|c|c|c|c|c}
\hline
Tokenizer & ace & ban & jav & min & sun & avg \\
\hline
Single-13L  &  \textbf{74.86 \textsubscript{($\pm$1.02)}} &  \textbf{77.01 \textsubscript{($\pm$0.80)}} &  \textbf{77.92 \textsubscript{($\pm$1.62)}} & 77.14 \textsubscript{($\pm$1.00)} &  \textbf{76.65 \textsubscript{($\pm$2.51)}} &  \textbf{76.72 \textsubscript{($\pm$1.39)}} \\
Parallel-13L (ours)  & 73.27 \textsubscript{($\pm$1.18)} & 73.76 \textsubscript{($\pm$0.43)} & 77.75 \textsubscript{($\pm$0.44)} &  \textbf{77.85 \textsubscript{($\pm$0.43)}} & 75.76 \textsubscript{($\pm$2.70)} & 75.68 \textsubscript{($\pm$1.04)} \\
\hline
\end{tabular}
\caption{NusaX-senti benchmarking results under monolingual setting. `ace', `ban', `jav', `min', and `sun' denote Acehnese, Balinese, Javanese, Minangkabau, and Sundanese, respectively.}
\label{tab:nusaxsenti_mono_scratch}
\end{table*}

\begin{table*}[tbp]
\centering
\tiny
\tabcolsep=0.13cm
\begin{tabular}{c|c|c|c|c|c|c|c|c|c}
\hline
Tokenizer & amh & hau & ibo & kin & orm & swh & tir & twi & avg \\
\hline
Single-13L  &  \textbf{60.92 \textsubscript{($\pm$2.41)}} & 74.39 \textsubscript{($\pm$0.65)} & 84.96 \textsubscript{($\pm$0.83)} &  \textbf{63.82 \textsubscript{($\pm$1.20)}} &  \textbf{73.60 \textsubscript{($\pm$1.29)}} & 90.13 \textsubscript{($\pm$0.18)} & 61.76 \textsubscript{($\pm$1.66)} &  \textbf{56.49 \textsubscript{($\pm$3.00)}} &  \textbf{62.17 \textsubscript{($\pm$1.75)}} \\
Parallel-13L (ours)  & 60.39 \textsubscript{($\pm$0.89)} &  \textbf{75.35 \textsubscript{($\pm$1.52)}} &  \textbf{87.70 \textsubscript{($\pm$0.26)}} & 62.94 \textsubscript{($\pm$1.54)} & 73.59 \textsubscript{($\pm$0.41)} &  \textbf{90.30 \textsubscript{($\pm$0.42)}} &  \textbf{61.94 \textsubscript{($\pm$2.77)}} & 55.20 \textsubscript{($\pm$1.15)} & 61.76 \textsubscript{($\pm$1.73)} \\
\hline
\end{tabular}
\caption{AfriHate benchmarking results under monolingual setting. `amh', `hau', `ibo', `kin', `orm', `swh', `tir', and `twi' denote Amharic, Hausa, Igbo, Kinyarwanda, Oromo, Swahili, Tigrinya, and Twi, respectively.}
\label{tab:afrihate_mono_scratch}
\end{table*}

\begin{table*}[tbp]
\centering
\begin{tabular}{c|c|c|c|c}
\hline
Tokenizer & amh & orm & tir & avg \\
\hline
Single-13L  & 58.70 \textsubscript{($\pm$1.32)} &  \textbf{60.15 \textsubscript{($\pm$0.75)}} & 48.92 \textsubscript{($\pm$1.24)} & 55.92 \textsubscript{($\pm$1.10)} \\
Parallel-13L (ours)  &  \textbf{60.98 \textsubscript{($\pm$1.09)}} & 59.63 \textsubscript{($\pm$1.63)} &  \textbf{52.58 \textsubscript{($\pm$0.20)}} &  \textbf{57.73 \textsubscript{($\pm$0.97)}} \\
\hline
\end{tabular}
\caption{EthioEmo benchmarking results under monolingual setting. `amh', `orm', and `tir' denote Amharic, Oromo, and Tigrinya, respectively.}
\label{tab:ethioemo_mono_scratch}
\end{table*}

\begin{table*}[tbp]
\centering
\small
\begin{tabular}{c|c|c|c|c|c|c}
\hline
Tokenizer & hau & ibo & kin & sun & swa & avg \\
\hline
Single-13L  & 57.90 \textsubscript{($\pm$1.40)} & 58.89 \textsubscript{($\pm$1.53)} & 43.89 \textsubscript{($\pm$0.20)} &  \textbf{61.05 \textsubscript{($\pm$0.93)}} & 23.81 \textsubscript{($\pm$1.14)} & 49.11 \textsubscript{($\pm$1.04)} \\
Parallel-13L (ours)  &  \textbf{60.23 \textsubscript{($\pm$0.35)}} &  \textbf{59.23 \textsubscript{($\pm$0.71)}} &  \textbf{47.74 \textsubscript{($\pm$2.45)}} & 60.97 \textsubscript{($\pm$0.79)} &  \textbf{24.04 \textsubscript{($\pm$2.15)}} &  \textbf{50.44 \textsubscript{($\pm$1.29)}} \\
\hline
\end{tabular}
\caption{BRIGHTER benchmarking results under monolingual setting. `hau', `ibo', `kin', `sun', and `swa',  denote Hausa, Igbo, Kinyarwanda, Sundanese, and Swahili, respectively.}
\label{tab:brighter_mono_scratch}
\end{table*}

\section{Continual Pre-Training: Representation Similarity}
\label{sec:apdx_cpt_representation_similarity}

\begin{figure}[t]
  \includegraphics[width=0.49\linewidth]{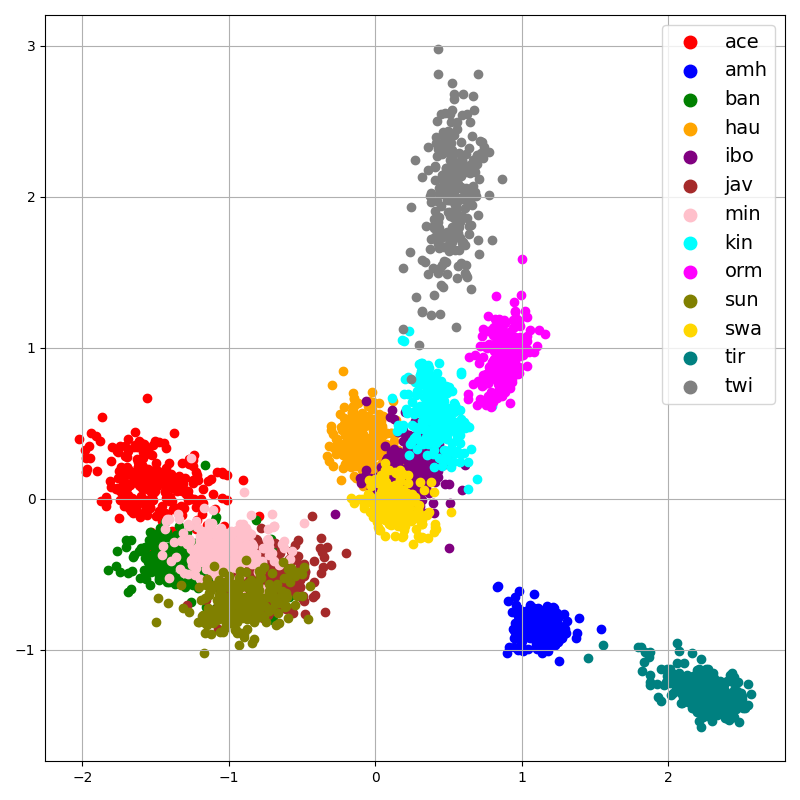} \hfill
  \includegraphics[width=0.49\linewidth]{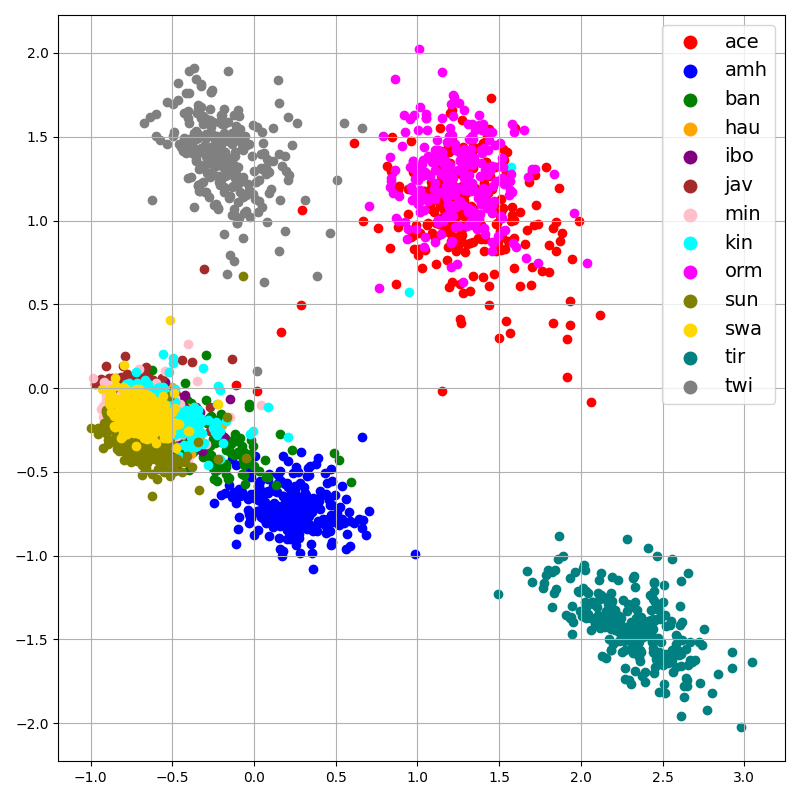}
  \caption {Principal Component Analysis (PCA) of continual pretrained model embeddings, comparing the \textit{Single-13L} (left) and the \textit{Parallel-13L} (right) on the parallel data FLORES+ dataset.}
  \label{fig:pca_flores_cpt}
\end{figure}

As discussed in Section~\ref{sec:cpt}, although the benchmarking performance of \textit{Single-13L} and \textit{Parallel-13L} after CPT is comparable, the cross-lingual representation similarity is stronger in \textit{Parallel-13L}. This is illustrated in Figure~\ref{fig:pca_flores_cpt}, where we visualize the embedding spaces of FLORES+~\citep{costa2022no} sentences for each language in two dimensions using PCA. In \textit{Single-13L}, inputs from the same language family remain clustered. For instance, Indonesian local languages (Acehnese, Minangkabau, Sundanese, etc.) cluster together, African languages cluster together, and Amharic-script languages (Amharic and Tigrinya) are separated into distinct clusters. In contrast, \textit{Parallel-13L} shows tighter cross-lingual clustering, with languages grouped more uniformly, including Amharic. The main outliers are Tigrinya and Twi, which remain separated in both models, likely due to limited pretraining resources.

These observations are further supported by our bitext mining analysis on the models after CPT. As shown in Table~\ref{tab:bitext_mining_cpt}, \textit{Parallel-13L} consistently achieves the lowest error rates and the highest number of best \texttt{xsim} scores, outperforming the other models, although the gains are smaller than those observed when pretraining from scratch. Tables~\ref{tab:cpt_bitext_mining_3} and \ref{tab:cpt_bitext_mining_5} provide the detailed \texttt{xsim} scores for each language pair in both \textit{Single-13L} and \textit{Parallel-13L}.

\begin{table*}
\centering
\small
\begin{tabular}{|l|c|c|c|c|c|c|c|c|c|c|c|c|c|}
\hline
 & ace & amh & ban & hau & ibo & jav & min & kin & orm & sun & swa & tir & twi \\
\hline
ace & \cellcolor{black!50} & \textbf{91.01} & \textbf{65.42} & 68.28 & \textbf{76.09} & \textbf{55.83} & \textbf{67.59} & \textbf{82.61} & \textbf{93.68} & 79.05 & 68.28 & \textbf{97.73} & \textbf{82.51} \\
\hline
amh & \cellcolor{black!50} & \cellcolor{black!50} & 93.48 & 90.42 & 89.72 & 88.24 & 90.51 & 91.90 & \textbf{94.96} & 89.92 & 90.81 & 92.29 & 94.76 \\
\hline
ban & \cellcolor{black!50} & \cellcolor{black!50} & \cellcolor{black!50} & \textbf{40.02} & \textbf{50.40} & \textbf{27.96} & \textbf{44.86} & 74.51 & 91.40 & 50.59 & 38.54 & 97.33 & 77.08 \\
\hline
hau & \cellcolor{black!50} & \cellcolor{black!50} & \cellcolor{black!50} & \cellcolor{black!50} & 36.26 & 29.25 & 41.80 & 67.00 & 90.61 & \textbf{40.32} & 24.90 & 95.55 & 69.76 \\
\hline
ibo & \cellcolor{black!50} & \cellcolor{black!50} & \cellcolor{black!50} & \cellcolor{black!50} & \cellcolor{black!50} & \textbf{43.68} & \textbf{55.14} & 74.11 & 92.59 & \textbf{53.56} & 40.51 & \textbf{96.44} & 76.38 \\
\hline
jav & \cellcolor{black!50} & \cellcolor{black!50} & \cellcolor{black!50} & \cellcolor{black!50} & \cellcolor{black!50} & \cellcolor{black!50} & 24.41 & 70.85 & 91.70 & \textbf{28.06} & 29.05 & 95.65 & 75.30 \\
\hline
min & \cellcolor{black!50} & \cellcolor{black!50} & \cellcolor{black!50} & \cellcolor{black!50} & \cellcolor{black!50} & \cellcolor{black!50} & \cellcolor{black!50} & \textbf{69.96} & 90.51 & \textbf{28.36} & 38.14 & 96.54 & 75.30 \\
\hline
kin & \cellcolor{black!50} & \cellcolor{black!50} & \cellcolor{black!50} & \cellcolor{black!50} & \cellcolor{black!50} & \cellcolor{black!50} & \cellcolor{black!50} & \cellcolor{black!50} & 94.37 & \textbf{73.62} & 66.21 & \textbf{96.84} & 85.57 \\
\hline
orm & \cellcolor{black!50} & \cellcolor{black!50} & \cellcolor{black!50} & \cellcolor{black!50} & \cellcolor{black!50} & \cellcolor{black!50} & \cellcolor{black!50} & \cellcolor{black!50} & \cellcolor{black!50} & \textbf{91.11} & 90.81 & \textbf{97.43} & \textbf{93.08} \\
\hline
sun & \cellcolor{black!50} & \cellcolor{black!50} & \cellcolor{black!50} & \cellcolor{black!50} & \cellcolor{black!50} & \cellcolor{black!50} & \cellcolor{black!50} & \cellcolor{black!50} & \cellcolor{black!50} & \cellcolor{black!50} & \textbf{33.10} & 96.05 & \textbf{75.79} \\
\hline
swa & \cellcolor{black!50} & \cellcolor{black!50} & \cellcolor{black!50} & \cellcolor{black!50} & \cellcolor{black!50} & \cellcolor{black!50} & \cellcolor{black!50} & \cellcolor{black!50} & \cellcolor{black!50} & \cellcolor{black!50} & \cellcolor{black!50} & 95.16 & 73.22 \\
\hline
tir & \cellcolor{black!50} & \cellcolor{black!50} & \cellcolor{black!50} & \cellcolor{black!50} & \cellcolor{black!50} & \cellcolor{black!50} & \cellcolor{black!50} & \cellcolor{black!50} & \cellcolor{black!50} & \cellcolor{black!50} & \cellcolor{black!50} & \cellcolor{black!50} & 98.02 \\
\hline
twi & \cellcolor{black!50} & \cellcolor{black!50} & \cellcolor{black!50} & \cellcolor{black!50} & \cellcolor{black!50} & \cellcolor{black!50} & \cellcolor{black!50} & \cellcolor{black!50} & \cellcolor{black!50} & \cellcolor{black!50} & \cellcolor{black!50} & \cellcolor{black!50} & \cellcolor{black!50} \\
\hline
\end{tabular}
\caption{Bitext mining metric scores on CPT models trained with \textit{Single-13L} tokenizer.}
\label{tab:cpt_bitext_mining_3}
\end{table*}

\begin{table*}
\centering
\small
\begin{tabular}{|l|c|c|c|c|c|c|c|c|c|c|c|c|c|}
\hline
 & ace & amh & ban & hau & ibo & jav & min & kin & orm & sun & swa & tir & twi \\
\hline
ace & \cellcolor{black!50} & 98.02 & 77.37 & \textbf{62.55} & 80.53 & 57.41 & 90.32 & 98.62 & 97.83 & \textbf{68.58} & \textbf{59.98} & 98.52 & 89.43 \\
\hline
amh & \cellcolor{black!50} & \cellcolor{black!50} & \textbf{74.90} & \textbf{62.65} & \textbf{78.46} & \textbf{63.54} & \textbf{64.62} & \textbf{76.88} & 95.06 & \textbf{71.15} & \textbf{59.49} & 97.33 & \textbf{86.07} \\
\hline
ban & \cellcolor{black!50} & \cellcolor{black!50} & \cellcolor{black!50} & 41.01 & 61.07 & 30.43 & 56.82 & \textbf{63.34} & \textbf{85.18} & \textbf{37.06} & \textbf{34.19} & \textbf{92.98} & \textbf{76.68} \\
\hline
hau & \cellcolor{black!50} & \cellcolor{black!50} & \cellcolor{black!50} & \cellcolor{black!50} & \textbf{31.82} & \textbf{20.06} & \textbf{28.75} & \textbf{47.92} & \textbf{79.15} & 53.56 & \textbf{17.39} & \textbf{90.61} & \textbf{63.24} \\
\hline
ibo & \cellcolor{black!50} & \cellcolor{black!50} & \cellcolor{black!50} & \cellcolor{black!50} & \cellcolor{black!50} & 43.97 & 69.86 & \textbf{66.60} & \textbf{85.18} & 62.35 & \textbf{36.76} & 97.04 & \textbf{72.63} \\
\hline
jav & \cellcolor{black!50} & \cellcolor{black!50} & \cellcolor{black!50} & \cellcolor{black!50} & \cellcolor{black!50} & \cellcolor{black!50} & \textbf{22.43} & \textbf{50.79} & \textbf{76.98} & 52.87 & \textbf{22.13} & \textbf{92.69} & \textbf{69.57} \\
\hline
min & \cellcolor{black!50} & \cellcolor{black!50} & \cellcolor{black!50} & \cellcolor{black!50} & \cellcolor{black!50} & \cellcolor{black!50} & \cellcolor{black!50} & 76.98 & \textbf{89.53} & 57.21 & \textbf{25.10} & \textbf{92.79} & \textbf{72.23} \\
\hline
kin & \cellcolor{black!50} & \cellcolor{black!50} & \cellcolor{black!50} & \cellcolor{black!50} & \cellcolor{black!50} & \cellcolor{black!50} & \cellcolor{black!50} & \cellcolor{black!50} & \textbf{91.70} & 83.60 & \textbf{54.64} & 97.04 & \textbf{78.56} \\
\hline
orm & \cellcolor{black!50} & \cellcolor{black!50} & \cellcolor{black!50} & \cellcolor{black!50} & \cellcolor{black!50} & \cellcolor{black!50} & \cellcolor{black!50} & \cellcolor{black!50} & \cellcolor{black!50} & 96.34 & \textbf{80.53} & 98.81 & 98.22 \\
\hline
sun & \cellcolor{black!50} & \cellcolor{black!50} & \cellcolor{black!50} & \cellcolor{black!50} & \cellcolor{black!50} & \cellcolor{black!50} & \cellcolor{black!50} & \cellcolor{black!50} & \cellcolor{black!50} & \cellcolor{black!50} & 40.32 & \textbf{90.12} & 85.28 \\
\hline
swa & \cellcolor{black!50} & \cellcolor{black!50} & \cellcolor{black!50} & \cellcolor{black!50} & \cellcolor{black!50} & \cellcolor{black!50} & \cellcolor{black!50} & \cellcolor{black!50} & \cellcolor{black!50} & \cellcolor{black!50} & \cellcolor{black!50} & \textbf{91.21} & \textbf{68.08} \\
\hline
tir & \cellcolor{black!50} & \cellcolor{black!50} & \cellcolor{black!50} & \cellcolor{black!50} & \cellcolor{black!50} & \cellcolor{black!50} & \cellcolor{black!50} & \cellcolor{black!50} & \cellcolor{black!50} & \cellcolor{black!50} & \cellcolor{black!50} & \cellcolor{black!50} & \textbf{97.92} \\
\hline
twi & \cellcolor{black!50} & \cellcolor{black!50} & \cellcolor{black!50} & \cellcolor{black!50} & \cellcolor{black!50} & \cellcolor{black!50} & \cellcolor{black!50} & \cellcolor{black!50} & \cellcolor{black!50} & \cellcolor{black!50} & \cellcolor{black!50} & \cellcolor{black!50} & \cellcolor{black!50} \\
\hline
\end{tabular}
\caption{Bitext mining metric scores on CPT models trained with \textit{Parallel-13L} tokenizer.}
\label{tab:cpt_bitext_mining_5}
\end{table*}

\begin{table}[ht]
\centering
\small
\tabcolsep=0.13cm
\begin{tabular}{l|cc}
\hline
& Single-13L & Parallel-13L \\
\hline
Avg. \texttt{xsim} $\downarrow$ & 72.18 & \textbf{69.34} \\ 
\# of best \texttt{xsim} $\uparrow$ & 29 & \textbf{48} \\
\hline
\end{tabular}
\caption{Bitext mining metric meta-scores on CPT models.}
\label{tab:bitext_mining_cpt}
\end{table}

\section{Model Input Representation Design}
\label{sec:proposed_method}

\begin{figure}[htbp]
  \includegraphics[width=0.99\linewidth]{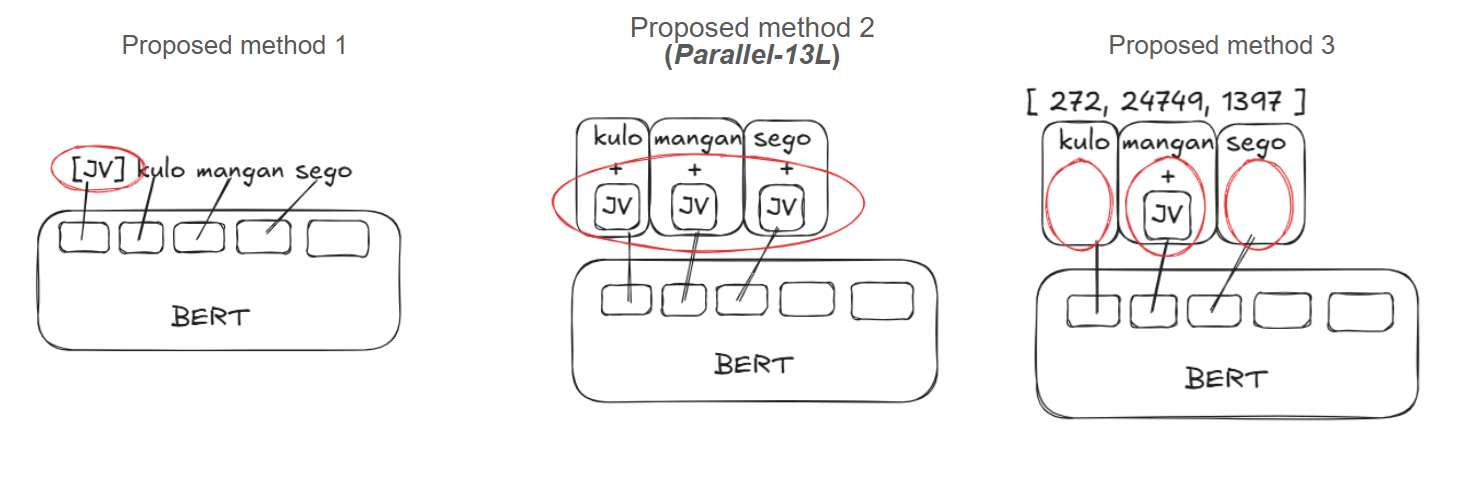}
  \caption {The design of the model input representation. We adopt the second method and refer to it as \textit{Parallel-13L}.}
  \label{fig:proposed_method}
\end{figure}

Since we require a `language token' to select the appropriate tokenizer for a given input, our input representation differs slightly from the conventional approach. During the model design phase, as presented in Figure~\ref{fig:proposed_method}, we explored several alternatives and ultimately selected the most effective method. Specifically, we considered three designs: (1) placing the language token at the beginning of every input sequence, requiring the model to process it as the first token; (2) the \textit{Parallel-13L} approach, where the language token serves as a signal that is added to each token embedding, effectively combining token and language embeddings; and (3) augmenting only the embeddings of unaligned vocabulary items (i.e., tokens not belonging to the \texttt{word}-type category\footnote{We store unaligned vocabulary items at the end of the vocabulary indices}).

We pretrained models with each of these methods from scratch using the same dataset (Wikipedia dumps for the 13 languages studied in this paper). The first and third approaches did not converge as quickly or effectively as \textit{Parallel-13L}, which therefore became our chosen design for the experiments reported in Section~\ref{sec:results_section} and beyond.

\end{document}